%% file: iclr2027_conference.tex
\documentclass{article}
\usepackage{iclr2027_conference,times}

\input{preamble}

\usepackage{hyperref}
\usepackage{url}
\usepackage{flafter}
\usepackage{placeins}
\usepackage[capitalize]{cleveref}

\crefname{section}{Sec.}{Secs.}
\Crefname{section}{Section}{Sections}
\crefname{subsection}{Sec.}{Secs.}
\Crefname{subsection}{Section}{Sections}
\crefname{table}{Tab.}{Tabs.}
\Crefname{table}{Table}{Tables}
\crefname{appendix}{Appendix}{Appendices}
\Crefname{appendix}{Appendix}{Appendices}

\title{ReSight-SMC: Two-Stage Power Sampling via Island SMC with Visual Scouts}
\author{\makebox[0.5\textwidth][l]{Yaowen Zhang} \\
Independent Researcher \\
\texttt{yaowenzhang196@outlook.com}
\And
\makebox[0.5\textwidth][l]{Xiangyu Qiu} \\
University of Electronic Science \\
and Technology of China \\
\texttt{xiangyuqiu@std.uestc.edu.cn}
\AND
\makebox[0.5\textwidth][l]{Junyi Hu} \\
Tsinghua University \\
\texttt{hujy24@mails.tsinghua.edu.cn}
\And
\makebox[0.5\textwidth][l]{Zhi Lu} \\
University of Electronic Science \\
and Technology of China \\
\texttt{zhilu@uestc.edu.cn}
\AND
\makebox[0.5\textwidth][l]{Wenwen Tian}\\
University of Electronic Science \\
and Technology of China \\
\texttt{wenwen\_tain@std.uestc.edu.cn}
\And
\makebox[0.5\textwidth][l]{Aoqin Wang} \\
University of Electronic Science \\
and Technology of China \\
\texttt{aoqin\_wang@std.uestc.edu.cn}
\AND
\makebox[0.5\textwidth][l]{Junhai Luo}\\
University of Electronic Science \\
and Technology of China \\
\texttt{junhai\_luo@uestc.edu.cn}
\And
\makebox[0.5\textwidth][l]{Zhenming Peng} \\
University of Electronic Science \\
and Technology of China \\
\texttt{zmpeng@uestc.edu.cn}
}
\iclrfinalcopy

\begin{document}
\maketitle
\lhead{Preprint}

\input{sec/0_abstract}
\input{sec/1_intro}
\input{sec/2_related}
\input{sec/4_method}

\input{sec/5_experiments}

\input{sec/6_discussion}

\input{sec/7_conclusion}



\bibliography{references}
\bibliographystyle{iclr2027_conference}

\clearpage
\appendix
\crefalias{section}{appendix}
\crefalias{subsection}{appendix}
\crefalias{subsubsection}{appendix}
\input{sec/A_analysis}
\input{sec/B_appendix}

\end{document}

%% file: preamble.tex
\usepackage{amsfonts,amssymb,mathtools}
\usepackage{amsthm}
\usepackage{microtype}
\usepackage{graphicx}
\usepackage{booktabs,multirow,tabularx,array}
\usepackage{algorithm}
\usepackage{algpseudocode}
\usepackage{tikz}
\usetikzlibrary{arrows.meta,positioning,fit,calc,backgrounds,shapes.geometric}
\usepackage{xcolor}
\usepackage{siunitx}
\usepackage{xspace}
\usepackage{makecell}
\definecolor{fullblue}{RGB}{45,92,156}
\definecolor{lightblue}{RGB}{222,235,249}
\definecolor{viewgreen}{RGB}{60,135,94}
\definecolor{lightgreen}{RGB}{223,242,230}
\definecolor{warmorange}{RGB}{194,102,39}
\definecolor{lightorange}{RGB}{250,231,213}
\definecolor{mutered}{RGB}{166,58,65}
\definecolor{lightred}{RGB}{247,224,226}
\newcommand{\method}{ReSight-SMC\xspace}
\newcommand{\ESS}{\operatorname{ESS}}

%% file: sec/0_abstract.tex
\begin{abstract}
Power sampling has emerged as a powerful training-free approach to LLM reasoning, eliciting capabilities comparable to reinforcement learning by sharpening the model distribution over complete responses. Despite this success, power sampling remains underexplored in large vision-language models (LVLMs). We first transfer Power-SMC to LVLM decoding by defining a sequence-power target over complete responses conditioned on both the image and the prompt.
This direct multimodal transfer provides a strong training-free
baseline, but leaves two aspects of finite-particle multimodal inference
unaddressed. At the particle level, global resampling can collapse genealogies, while particle-based power sampling does not diversify trajectories through distinct visual cues in multimodal decoding, limiting exploration under a finite particle budget. At the answer level, sequence-level
sharpening makes distinct reasoning trajectories compete even when they support
the same answer. We introduce \method, a verifier-free two-stage power sampler for LVLM inference. Its first stage uses ancestry-isolated SMC islands to preserve independent trajectory families and routes a bounded set of prefix-conditioned visual scouts to prefix-relevant image regions while discouraging redundant overlap. Each scout temporarily increases attention to the image tokens and further emphasizes its routed region. Exact importance correction preserves the base LVLM sequence-power target. The second stage aggregates terminal importance mass by canonical answer, powers the resulting answer marginal, and samples an answer together with a supporting trajectory. Across four LVLM backbones and five benchmarks, \method achieves stronger aggregate performance than Power-SMC over both the reasoning and perception benchmark groups. Without post-training, it remains competitive in aggregate with backbone-matched models trained using reinforcement learning.
Code is available at \url{https://github.com/yaowenzhang1/resight-smc}. 
\end{abstract}

%% file: sec/1_intro.tex
\section{Introduction}
\label{sec:intro}
Test-time scaling improves LLM reasoning without parameter updates by allocating
additional inference compute to sampling, aggregation, search, or refinement
~\citep{wei2022cot,wang2023selfconsistency,sammani2026tts}. Sequence-level power sampling provides a model-internal target for test-time
scaling. Given a base distribution $p_\theta(y\mid x)$, it samples complete
responses from the sharpened distribution
$\pi_\alpha(y\mid x)\propto p_\theta(y\mid x)^\alpha$, where $\alpha>1$. This sharpens complete-response
probabilities rather than independently lowering each next-token temperature.
Reasoning with Sampling~\citep{karan2025reasoning} shows that this target can
approach reinforcement-learning performance without training or external
verifiers, while Power-SMC~\citep{azizi2026powersmc} implements it efficiently
with parallel weighted trajectories. Despite its success in LLMs, sequence-level power sampling remains
underexplored for open-ended LVLM decoding. Existing multimodal applications focus on embodied control or iterative visual-reasoning refinement
\citep{park2026policy,chen2026drift,jiang2026aligning}. We transfer
Power-SMC~\citep{azizi2026powersmc} to LVLM decoding by defining its target over
complete responses conditioned jointly on the image and prompt. As our
experiments show, this direct multimodal transfer already provides a strong
training-free baseline.

However, global resampling can repeatedly duplicate high-weight trajectories,
reducing genealogical coverage and causing the classical path-degeneracy problem in SMC~\citep{delmoral2006smc,doucet2011tutorial}. During long visual reasoning,
visual reliance can decline as the textual trace grows
\citep{sammani2026tts,favero2024m3id}, while direct Power-SMC lacks a mechanism for leveraging visual attention early in generation to promote trajectory diversity. Moreover, sequence-level power
sharpens individual trajectories before answer aggregation. Trajectories that
support the same answer therefore compete through their separate sequence-level weights,
allowing one or a few high-probability paths to outweigh collective support
distributed across multiple moderate-probability paths
\citep{yang2026correctmass}.

We introduce \method, a verifier-free two-stage power sampler for LVLM
inference. Its first stage improves finite-particle coverage of the base LVLM
sequence-power target. Ancestry-isolated islands resample independently to
preserve distinct trajectory families. At a prescribed checkpoint, current prefixes route a bounded set of visual scouts to image regions in the input image while discouraging redundant spatial overlap. 
Each scout temporarily reactivates attention to all image tokens and gives its assigned region an additional boost. Unfinished base-proposal anchors remain active under the original proposal. Exact importance correction
preserves the base LVLM target, so scouting changes exploration
rather than the distribution being estimated.

The second stage aggregates normalized terminal trajectory mass by canonical
answer, raises the resulting answer masses to a finite power $\gamma$, and
samples an answer together with a supporting trajectory. Applying the second
power after aggregation balances a few high-weight trajectories against broader
same-answer support. Across four LVLM backbones and five benchmarks spanning
mathematical, logical, and general multimodal reasoning, perception, and
real-world spatial understanding, \method outperforms Power-SMC in aggregate on
both reasoning and perception-focused tasks. Without post-training, it remains
competitive with backbone-matched TRACE-RL and surpasses Game-RL across all benchmarks at the 7B scale.

Our contributions are:
\begin{itemize}
\item We extend Power-SMC~\citep{azizi2026powersmc}, which approximates the
sequence-power target with parallel weighted particles, to open-ended LVLM
reasoning.

\item We formulate two-stage sequence-to-answer power sampling, combining a
sequence-power population with answer-marginal sharpening.

\item We formulate an autoregressive island SMC sampler that combines
ancestry-isolated resampling with exact sequential importance correction for
the sequence-power target.

\item We design prefix-conditioned visual scouts that reactivate attention to routed image regions to diversify finite-particle exploration.
\end{itemize}

%% file: sec/2_related.tex
\section{Related work}
\label{sec:related}

\paragraph{LVLM test-time scaling.}
LVLM test-time scaling spans structured multimodal reasoning and the transfer of sampling, aggregation, and refinement strategies
~\citep{mitra2024ccot,sammani2026tts}.
Longer traces can improve difficult problems but also weaken attention to image
tokens~\citep{sammani2026tts}. AVIS~\citep{jeddi2026avis} jointly adapts
visual-token retention and the reasoning-rollout budget through key-based
pruning and a learned difficulty predictor. TTAdapt~\citep{kaya2025efficient}
updates model parameters at inference from consensus pseudolabels.
\method keeps its parameters fixed and introduces scout proposals within a verifier-free weighted sampler.

\paragraph{Sequence-level power sampling and island SMC.}
Reasoning with Sampling~\citep{karan2025reasoning} targets
$p(y\,|\,x)^\alpha$ with suffix-resampling Metropolis--Hastings (MH). Later
work improves cut selection and scalability and connects power distributions
to self-reward objectives
~\citep{zhou2026decision,ji2026scalable,tomihari2026bridges}.
MH is serial and requires $16\text{--}28\times$ the latency of standard
decoding on MATH500. Power-SMC~\citep{azizi2026powersmc} instead advances weighted particles in
parallel and resamples when the ESS falls below a threshold, reducing
MATH500 latency to $1.44$--$3.25\times$ standard decoding across four models. Global resampling can duplicate a few high-weight
trajectories, a classical particle-impoverishment failure mode
~\citep{delmoral2006smc}. We use an island particle system
~\citep{verge2015island} that confines resampling ancestry within each island.
Island parallelism also appears in SMCEvolve~\citep{jiang2026smcevolve}, which
uses migratory SMC chains for reward-tilted program search, whereas our
non-migrating islands approximate a common sequence-power target in token-level
autoregressive decoding. Multimodal power sampling has so far focused on embodied action and planning
\citep{park2026policy,chen2026drift} or MCMC-based refinement of open-ended
LVLM reasoning~\citep{jiang2026aligning}. We transfer Power-SMC to open-ended
LVLM decoding under a base LVLM sequence-power target and extend it with
ancestry-isolated islands and prefix-conditioned visual proposals.

\paragraph{Answer aggregation and marginal sharpening.}
Self-consistency~\citep{wang2023selfconsistency} samples multiple reasoning paths
and returns the most frequent answer as a deterministic decision rule. Marginal
sharpening~\citep{arzhantsev2026marginal} instead defines a powered answer
marginal and approximately samples from it through multi-trace decoding. Our
readout starts from a sequence-power population: it aggregates corrected SMC
mass by canonical answer, applies a second finite power, and samples from the
resulting distribution. This composes sequence- and answer-level sharpening
without unweighted voting. \citet{yang2026correctmass} modify the trajectory target to preserve support
and control per-problem deformation. Our second-stage readout leaves the
sequence target unchanged and applies finite-power sharpening after aggregating
trajectory mass by answer.


\paragraph{Visual conditioning during inference.}
Training-free methods modulate visual influence through image-signal
amplification~\citep{favero2024m3id}, contrastive decoding
~\citep{leng2024vcd}, attention-driven region selection
~\citep{gao2025icot}, or external visual tools
~\citep{wang2025visuothink}. \method introduces prefix-conditioned visual
scouts to diversify particle proposals: each reactivates attention to all image tokens and
emphasizes a routed region, while exact importance correction preserves the
base LVLM sequence-power target.

%% file: sec/4_method.tex
\section{Method}
\label{sec:method}

\subsection{Overview}
\method separates sharpening into a trajectory stage and a readout stage.
The trajectory stage uses island SMC to construct a weighted population for a
base LVLM sequence-power target. Islands resample independently so that a
single successful lineage cannot eliminate all genealogical diversity. At a
prespecified checkpoint, a bounded subset of particles becomes visual scouts:
their current prefixes guide routing among candidate image regions under an
overlap penalty, and each scout temporarily samples from a proposal that increases
attention to image tokens and further emphasizes its routed region. Exact importance correction preserves the same base LVLM sequence-power target.

After generation, the readout stage pools terminal particle mass by canonical answer
and applies a finite power to this answer marginal. This lets an answer draw strength from support distributed across multiple moderate-weight trajectories. By contrast, a single-stage sequence-power sampler sharpens trajectories individually and can concentrate mass on a few dominant paths, even when other paths support the same answer. 

\begin{figure*}[t]
\centering
\includegraphics[width=0.98\textwidth]{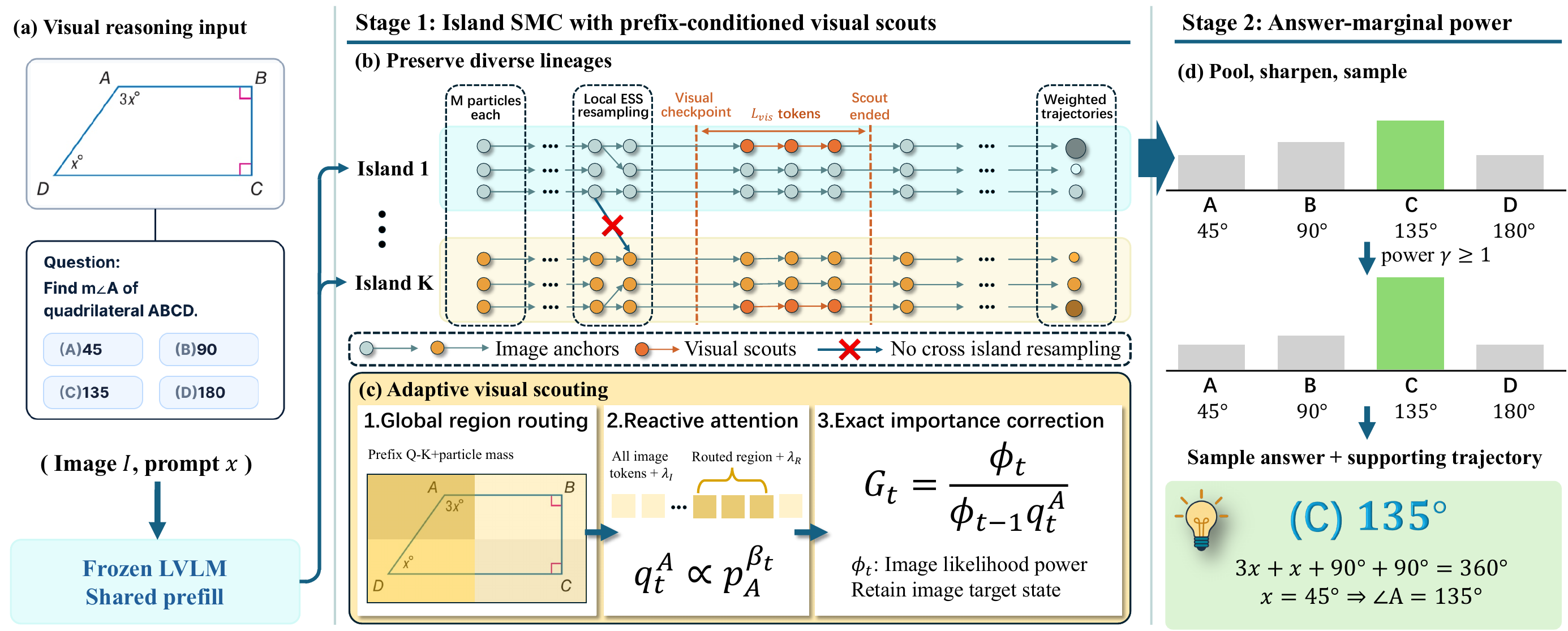}
\caption{\textbf{ReSight-SMC.}
Stage one maintains ancestry-isolated SMC islands for the base LVLM
sequence-power target. Islands test for ESS-triggered resampling at regular
token intervals. At a prespecified visual checkpoint, the remaining unfinished particles
continue with the base proposal, while selected scouts use prefix-conditioned
routing and attention-reactivated image tokens to form scout
proposals. Exact
importance correction preserves the base LVLM target. Stage two aggregates
terminal mass by answer, applies a finite answer power, and samples an answer
with a supporting trajectory.}
\label{fig:method}
\end{figure*}

Together, these components form a coverage-to-decision pipeline. Island-local
resampling preserves distinct lineages, so visual scouts operate over more
diverse reasoning prefixes. The scouts can discover region-guided continuations that the base proposal may miss in a finite population. Importance correction assigns them substantial weight only when they are also supported by the base LVLM sequence-power target. The answer readout then pools corrected mass across trajectories supporting the same answer and sharpens the resulting answer marginal, allowing these trajectories to contribute jointly to answer selection. \Cref{fig:method} summarizes the pipeline, and
\cref{app:algorithm} gives the complete inference procedure.

\subsection{LVLM sequence-power target and island SMC}
Let $I$ be an image, $x$ a prompt, $H$ the decoding horizon, and
$y=(y_1,\ldots,y_T)$ a response completed at step $T\leq H$. The fixed LVLM
distribution factorizes as
$p_F(y\mid I,x)=\prod_{s=1}^T p_F(y_s\mid y_{<s},I,x)$, and stage one targets
\begin{equation}
\pi_\alpha^F(y\mid I,x)
=\frac{p_F(y\mid I,x)^\alpha}{Z_\alpha^F(I,x)},
\qquad \alpha>1.
\label{eq:finite_power_target}
\end{equation}
We apply Power-SMC~\citep{azizi2026powersmc} to the autoregressive response tokens while leaving the model's native image conditioning unchanged. It approaches the LVLM sequence-power target through intermediate exponents
$1=\beta_0\leq\cdots\leq\beta_H=\alpha$. At step $t$, the unnormalized
target is
$\varphi_t(y_{1:t})=p_{F,t}(y_{1:t}\mid I,x)^{\beta_t}$ and the base
proposal locally powers the native next-token conditional,
$q_t^F(v\mid y_{<t},I,x)\propto
p_F(v\mid y_{<t},I,x)^{\beta_t}$. 
Local normalization depends on the prefix, so these conditionals alone do not induce \cref{eq:finite_power_target}. The generic importance weight in \cref{eq:generic_increment} supplies the required sequence-level correction. Its expansion and terminal-sequence treatment are derived in \cref{app:weights}.

We maintain $K$ isolated islands with $M$ particles each. A single multimodal prefill of $(I,x)$ produces an image-conditioned decoder state from which all $KM$ particles are initialized. Let
$\mathcal F_{t-1}$ denote the particle-system history available before step
$t$, including particle prefixes, weights, ancestry, completion states, and any
active scout assignments. At step $t$, the sampler uses $\mathcal F_{t-1}$ to select the decoder branch and its settings.
For particle $(k,m)$, the selected branch runs the LVLM autoregressively on
$(I,x,y_{<t}^{k,m})$ and produces next-token logits. Their softmax is the branch-specific
next-token distribution. At step $t$, particle $(k,m)$ draws its next token from the locally powered proposal
\begin{equation}
q_t^{F,k,m}(v\mid y_{<t}^{k,m},I,x)
=
\frac{p_F(v\mid y_{<t}^{k,m},I,x)^{\beta_t}}
{\sum_{u\in\mathcal V}p_F(u\mid y_{<t}^{k,m},I,x)^{\beta_t}}.
\label{eq:full_image_proposal}
\end{equation}
The next token is drawn as $y_t^{k,m}\sim q_t^{k,m}$. The incremental importance ratio below corrects this branch-specific proposal toward the intermediate target:
\begin{equation}
\begin{aligned}
G_t^{k,m}
&=\frac{\varphi_t(y_{1:t}^{k,m})}
{\varphi_{t-1}(y_{<t}^{k,m})
q_t^{k,m}(y_t^{k,m}\mid y_{<t}^{k,m},\mathcal F_{t-1})},\\
w_0^{k,m}&=\frac{1}{M},
\qquad w_t^{k,m}=w_{t-1}^{k,m}G_t^{k,m}.
\end{aligned}
\label{eq:generic_increment}
\end{equation}
Here $G_t^{k,m}$ is the incremental correction for the sampled token, and $w_t^{k,m}$ is the resulting unnormalized particle weight. A particle that emits EOS remains fixed while the remaining bridge corrections bring its completed response to the common terminal exponent $\alpha$. Within island \(k\), the weights are normalized as
\(\bar w_t^{k,m}=w_t^{k,m}/\sum_{j=1}^{M}w_t^{k,j}\).
The corresponding effective sample size is
\(\ESS_{k,t}=(\sum_{m=1}^{M}(\bar w_t^{k,m})^{\,2})^{-1}\). We evaluate this criterion every \(L_{\mathrm{SMC}}\) generated tokens. At each checkpoint, island \(k\) resamples only when
\(\ESS_{k,t}<\rho_wM\). When triggered, island \(k\) independently applies stratified resampling:
\begin{equation}
A_{k,t}^{1:M}
\sim \operatorname{StratResample}(\bar w_t^{k,1:M}),
\qquad A_{k,t}^m\in\{1,\ldots,M\}.
\label{eq:local_resampling}
\end{equation}
Here, \(A_{k,t}^m\) is the within-island ancestor selected for descendant
\(m\). Power-SMC~\citep{azizi2026powersmc} uses systematic resampling. We instead use stratified resampling because it is not only conditionally
unbiased but also consistent as $M\to\infty$ without imposing an ordering
condition on the particles~\citep{gerber2019negative}. Descendants receive weight \(1/M\),
and all ancestors remain within island \(k\), preventing population-wide
genealogical collapse.

Before any resampling at step \(t\), each island updates its normalizing-constant estimate as
\begin{equation}
c_{k,t}=\sum_{m=1}^M\bar w_{t-1}^{k,m}G_t^{k,m},
\qquad
\widehat Z_{k,t}=\widehat Z_{k,t-1}c_{k,t},
\qquad \widehat Z_{k,0}=1.
\label{eq:island_mass}
\end{equation}
The terminal-sequence convention and derivation of these recursions are given
in \cref{app:island_mechanics}.

At termination, island mass and normalized within-island weight combine into
the pooled first-stage particle mass
\begin{equation}
\widetilde W_H^{k,m}
=\frac{\widehat Z_{k,H}\bar w_H^{k,m}}
{\sum_{j=1}^K\widehat Z_{j,H}}.
\label{eq:island_terminal_weight}
\end{equation}
Because particle weights are normalized separately within each island, \(\bar w_H^{k,m}\) only compares particles within island \(k\). Multiplying by \(\widehat Z_{k,H}\) yields pooled particle masses used by the answer-level readout.

\subsection{Prefix-conditioned visual scouts}
The visual-scout episode is a bounded intervention invoked once at the prescribed checkpoint
$\tau_{\mathrm{vis}}$. Each eligible island receives a scout quota of $\lceil\rho_VM\rceil$, capped
to retain at least one unfinished base-proposal
anchor. A population router pairs these scouts with regions from a fixed multiscale grid
bank $\mathcal R=\{R_1,\ldots,R_G\}$. Each scout emphasizes its assigned
region for $L_{\mathrm{vis}}$ tokens and resumes the base proposal before the next
resampling checkpoint. This introduces region-guided proposal variation across selected particles while retaining base-proposal anchors.

\paragraph{Routing scouts to evidence.}
At $\tau_{\mathrm{vis}}$, we compute a visual-only Q--K softmax for the current prefix,
normalizing separately within each head over the original image tokens. For each
candidate region, we sum this relevance over its tokens, average across heads,
and apply region-size normalization:
\begin{equation}
A_g^{k,m}
=\frac{1}{N_h|R_g|^\zeta}
\sum_{a=1}^{N_h}\sum_{i\in R_g}\xi_{i,a}^{k,m},
\qquad 0\leq\zeta\leq1.
\label{eq:particle_region_score}
\end{equation}
Here $\xi_{i,a}^{k,m}$ is the visual-only Q--K softmax weight assigned by
attention head $a$ at the final decoder layer to image token $i$ for particle
$(k,m)$, normalized over all original-image tokens. $N_h$ is the
number of attention heads, $|R_g|$ is the number of visual tokens covered by
region $R_g$, and $\zeta$ controls the degree of region-size normalization. We combine this score with the particle's pooled SMC mass
$\widetilde W_{\tau_{\mathrm{vis}}}^{k,m}$ to define the
particle--region utility:
\begin{equation}
u_{k,m,g}
=\widetilde W_{\tau_{\mathrm{vis}}}^{k,m}
\frac{A_g^{k,m}}{\max(\sum_{g'=1}^G A_{g'}^{k,m},\epsilon)}.
\label{eq:particle_region_utility}
\end{equation}
Here $\epsilon=10^{-12}$ prevents division by zero when a particle assigns
zero total relevance to the candidate region bank. We min--max normalize these utilities over the eligible particle--region candidates to obtain $\widetilde u_{k,m,g}\in[0,1]$. The router then greedily selects
\begin{equation}
(k_b,m_b,g_b)
=\arg\max_{(k,m,g)\in\mathcal C_b}
\left[\widetilde u_{k,m,g}
-\mu\max_{r<b}\operatorname{IoU}(R_g,R_{g_r})\right].
\label{eq:greedy_route}
\end{equation}
Exact score ties favor larger $k$, then $m$, and finally $g$. The candidate set $\mathcal C_b$ enforces the per-island quota and assigns at
most one region to each particle. The utility directs scouts toward
prefix-relevant evidence carried by high-mass hypotheses. Here $\mu$ is the global
soft-overlap coefficient; it discourages redundant assignments without forbidding overlapping evidence. Exact Q--K
aggregation, mass and utility normalization, quota edge cases, and routing
complexity are given in \cref{app:routing_spec}.

\paragraph{Attention reactivation and proposal correction.}
Let $\mathcal V_I$ denote the image-token positions and
$\mathcal V_I(R_g)\subseteq\mathcal V_I$ the positions covered by a routed
region. For a scout assigned to $R_g$, we add the following bias to the causal
self-attention logits:
\begin{equation}
b_j(R_g)=\lambda_I\mathbf 1[j\in\mathcal V_I]
+\lambda_R\mathbf 1[j\in\mathcal V_I(R_g)].
\label{eq:attention_bias}
\end{equation}
At every decoder layer, the same visual-token logit bias is added to the current query's attention logits across all heads. It raises attention to all image tokens while giving the routed region an
additional boost. Let $p_A^{k,m}$ be the frozen
LVLM conditional under this intervention. The scout samples directly from
its locally powered proposal
\begin{equation}
q_t^{A,k,m}(v)
=\frac{p_A^{k,m}(v\mid y_{<t},I,x)^{\beta_t}}
{\sum_u p_A^{k,m}(u\mid y_{<t},I,x)^{\beta_t}}.
\label{eq:attention_proposal}
\end{equation}
To make the intervention effective from the scout's first newly generated token, we fork the selected particle's decoder state, roll back its cached latest-token computation, and recompute that fixed token once under \cref{eq:attention_bias}. This updates the next-token proposal without duplicating or resampling any response token. During the bounded episode, every token sampled from the scout proposal is also passed through an unmodified persistent base target state. This parallel state provides the base-model likelihood required for importance correction and maintains a bias-free KV cache along the same sampled trajectory. The biased state is advanced only when another scout proposal is required and is discarded after \(L_{\mathrm{vis}}\) tokens, after which decoding continues from the persistent base target state.

The scout branch changes only the proposal. Its exact probability enters
the incremental weight:
\begin{equation}
G_t^{A,k,m}
=\frac{\varphi_t(y_{1:t}^{k,m})}
{\varphi_{t-1}(y_{<t}^{k,m})
q_t^{A,k,m}(y_t^{k,m})}.
\label{eq:visual_increment}
\end{equation}
Since $\varphi_t$ is evaluated under the base LVLM, this ratio supplies exact importance correction for
every scout transition to the sequence target in
\cref{eq:finite_power_target}. Routing therefore alters finite-population
exploration and subsequent ancestry while maintaining proper weighting for the same sequence target. The importance-weight derivation is given in \cref{app:weights}, and the
finite-sample proper-weighting argument is given in
\cref{app:routingproof}.

\subsection{Answer-marginal power readout}
\label{sec:answer_readout}
Let $\operatorname{Ans}(y)$ be the canonical answer of response $y$. We first pool the
terminal masses in \cref{eq:island_terminal_weight} by answer and then apply a
finite answer-level power:
\begin{align}
\widehat\mu_\alpha(a)
&=\sum_{k=1}^K\sum_{m=1}^M
\widetilde W_H^{k,m}\mathbf 1[\operatorname{Ans}(y^{k,m})=a],
\label{eq:empirical_answer_marginal}\\
\widehat q_{\alpha,\gamma}(a)
&=\frac{\widehat\mu_\alpha(a)^\gamma}
{\sum_b\widehat\mu_\alpha(b)^\gamma},
\qquad \gamma\geq1.
\label{eq:empirical_answer_power}
\end{align}
We draw the final answer from $\widehat q_{\alpha,\gamma}$.
At the population level, $(\alpha,\gamma)=(2,2)$ produces a fourth-order score in the base trajectory probabilities. Unlike
direct $\alpha=4$ sequence sampling, the second square is applied
after trajectories supporting the same answer have pooled their mass. It
therefore preserves cross-trajectory support, reducing domination by a few
high-weight paths without reverting to unweighted majority voting. A formal
comparison, a separating example, the optional recovery of a supporting
trajectory, and the corresponding population-level output distribution are given in
\cref{app:split_sharpening,app:answer_power}.

%% file: sec/5_experiments.tex
\section{Experiments}
\label{sec:experiments}

\subsection{Experimental setup}
Our experiments examine whether sequence-power sampling is effective when
transferred to open-ended LVLM decoding and whether \method further improves
this direct-transfer baseline. We evaluate Qwen2.5-VL-3B-Instruct, Qwen2.5-VL-7B-Instruct,
Qwen3-VL-4B-Instruct, and Qwen3-VL-8B-Instruct
~\citep{bai2025qwen25vl,bai2025qwen3vl}. The benchmarks are MathVista, LogicVista, MMStar, and RealWorldQA
\citep{lu2023mathvista,xiao2024logicvista,chen2024mmstar,xai2024realworldqa}.
We partition MMStar into reasoning (MMStar-R) and perception (MMStar-P)
subsets (see \cref{app:prompt_eval} for details). Together, these benchmarks cover visual mathematical reasoning,
visual logical reasoning, broad perception, and real-world spatial
understanding. Accuracy is measured on each
released evaluation split.

All systems use the same image preprocessing, answer evaluator, response
horizon, and task-specific prompt for a given backbone. MathVista, MMStar-R, and
LogicVista use chain-of-thought prompting. MMStar-P and RealWorldQA use direct
answering. We evaluate the ESS criterion every 32 generated tokens and resample an island only when its ESS falls below the threshold. Visual scouting is invoked
once at $\tau_{\mathrm{vis}}=40$ and is skipped when generation terminates earlier. Population methods
use 32 particles. We report pass@1 as the mean $\pm$ sample standard
deviation over four runs with random seeds $\{0,1,2,3\}$.
Complete hyperparameters, benchmark protocols, and numerical implementation details are
provided in \cref{app:additional}.

\paragraph{Compared systems.}
For every backbone, we compare the base model, low-temperature autoregressive
sampling, our direct LVLM transfer of Power-SMC~\citep{azizi2026powersmc}, and
\method. The transferred Power-SMC baseline applies sequence-power SMC to
autoregressive response tokens conditioned on the backbone's native full-image
prefill and uses the same backbone, prompt, and total particle budget as
\method. All sharpening baselines use the same
sequence-level exponent $\alpha=2$. Low-temperature sampling uses
temperature $1/\alpha=0.5$. \method uses $K=4$ islands with $M=8$ particles each and answer exponent $\gamma=2$. 

At the Qwen2.5-VL scales, we additionally evaluate the released TRACE-RL 3B and
7B checkpoints, each trained with GRPO from the corresponding Instruct backbone
on procedural visual-reasoning data~\citep{alam2026trace}. At 7B we include
Game-RL-Qwen2.5-VL-7B, which applies GRPO on GameQA
~\citep{tong2025code2logic}. These rows provide backbone-matched post-training
references.

\subsection{Results across backbones}
\Cref{tab:main_results} compares all systems across backbones and benchmarks.
\begin{table*}[!htbp]
\caption{Pass@1 accuracy (\%) across backbones and benchmarks, reported as
mean $\pm$ sample standard deviation over four seeds. The all-data average is
question-weighted. Bold denotes the best non-RL result within each backbone.}
\label{tab:main_results}
\begin{center}
\scriptsize
\setlength{\tabcolsep}{4.5pt}
\begin{tabular}{lccccc|c}
\toprule
System & LogicVista & MathVista & MMStar-R & MMStar-P & RealWorldQA & All-data avg.\\
\midrule
\multicolumn{6}{l|}{\textit{Qwen2.5-VL-3B-Instruct}} & \\
Base & \(31.8 \pm 2.7\) & \(46.2 \pm 1.8\) & \(42.9 \pm 0.5\) & \(49.8 \pm 4.2\) & \(53.4 \pm 4.7\) & \(45.5 \pm 1.8\)\\
Low-temp. sampling & \(33.5 \pm 1.2\) & \(54.4 \pm 1.0\) & \(\mathbf{48.3 \pm 2.1}\) & \(54.9 \pm 1.8\) & \(58.2 \pm 0.9\) & \(51.1 \pm 0.4\)\\
Power-SMC & \(34.9 \pm 1.1\) & \(\mathbf{54.6 \pm 0.6}\) & \(47.0 \pm 1.7\) & \(53.5 \pm 1.5\) & \(56.1 \pm 1.5\) & \(50.3 \pm 0.5\)\\
\method (Ours) & \(\mathbf{35.8 \pm 1.1}\) & \(54.2 \pm 1.1\) & \(47.2 \pm 1.2\) & \(\mathbf{56.3 \pm 1.2}\) & \(\mathbf{58.8 \pm 0.5}\) & \(\mathbf{51.3 \pm 0.5}\)\\
TRACE-RL & \(37.9 \pm 2.7\) & \(56.4 \pm 0.8\) & \(53.7 \pm 0.4\) & \(53.2 \pm 3.1\) & \(55.5 \pm 1.9\) & \(52.8 \pm 0.8\)\\
\midrule
\multicolumn{6}{l|}{\textit{Qwen2.5-VL-7B-Instruct}} & \\
Base & \(40.3 \pm 0.9\) & \(66.5 \pm 0.3\) & \(61.0 \pm 1.0\) & \(60.2 \pm 1.3\) & \(66.0 \pm 2.5\) & \(60.9 \pm 0.5\)\\
Low-temp. sampling & \(41.2 \pm 1.3\) & \(69.5 \pm 0.3\) & \(62.6 \pm 1.4\) & \(61.5 \pm 1.0\) & \(69.5 \pm 1.9\) & \(63.1 \pm 0.3\)\\
Power-SMC & \(42.8 \pm 1.0\) & \(70.2 \pm 0.7\) & \(63.7 \pm 0.8\) & \(61.7 \pm 0.9\) & \(69.2 \pm 0.4\) & \(63.8 \pm 0.4\)\\
\method (Ours) & \(\mathbf{43.5 \pm 2.1}\) & \(\mathbf{71.7 \pm 0.8}\) & \(\mathbf{65.3 \pm 1.0}\) & \(\mathbf{63.1 \pm 0.4}\) & \(\mathbf{70.0 \pm 0.7}\) & \(\mathbf{65.0 \pm 0.3}\)\\
TRACE-RL & \(44.0 \pm 1.2\) & \(74.3 \pm 1.5\) & \(66.6 \pm 0.8\) & \(62.6 \pm 0.0\) & \(67.6 \pm 1.8\) & \(65.6 \pm 0.6\)\\
Game-RL & \(41.4 \pm 3.3\) & \(66.4 \pm 1.2\) & \(61.1 \pm 0.4\) & \(61.0 \pm 1.9\) & \(66.1 \pm 2.4\) & \(61.1 \pm 0.2\)\\
\midrule
\multicolumn{6}{l|}{\textit{Qwen3-VL-4B-Instruct}} & \\
Base & \(33.9 \pm 1.4\) & \(62.7 \pm 0.7\) & \(57.0 \pm 0.6\) & \(64.0 \pm 1.8\) & \(69.0 \pm 1.1\) & \(59.2 \pm 0.4\)\\
Low-temp. sampling & \(36.2 \pm 1.0\) & \(63.5 \pm 0.9\) & \(57.3 \pm 0.9\) & \(64.5 \pm 0.5\) & \(68.9 \pm 0.6\) & \(59.8 \pm 0.3\)\\
Power-SMC & \(45.2 \pm 1.0\) & \(72.5 \pm 0.6\) & \(65.4 \pm 0.9\) & \(67.7 \pm 0.8\) & \(69.8 \pm 0.5\) & \(66.1 \pm 0.5\)\\
\method (Ours) & \(\mathbf{48.8 \pm 0.6}\) & \(\mathbf{73.6 \pm 0.4}\) & \(\mathbf{67.9 \pm 0.8}\) & \(\mathbf{68.2 \pm 0.7}\) & \(\mathbf{70.8 \pm 0.6}\) & \(\mathbf{67.8 \pm 0.2}\)\\
\midrule
\multicolumn{6}{l|}{\textit{Qwen3-VL-8B-Instruct}} & \\
Base & \(40.0 \pm 0.9\) & \(66.1 \pm 1.6\) & \(60.8 \pm 0.6\) & \(65.5 \pm 1.5\) & \(69.9 \pm 1.4\) & \(62.2 \pm 0.4\)\\
Low-temp. sampling & \(43.2 \pm 1.0\) & \(66.9 \pm 1.0\) & \(60.6 \pm 0.6\) & \(63.0 \pm 1.8\) & \(70.1 \pm 0.9\) & \(62.5 \pm 0.4\)\\
Power-SMC & \(49.8 \pm 1.0\) & \(74.8 \pm 0.7\) & \(68.1 \pm 0.5\) & \(67.7 \pm 0.5\) & \(\mathbf{70.5 \pm 0.6}\) & \(68.1 \pm 0.2\)\\
\method (Ours) & \(\mathbf{53.3 \pm 1.1}\) & \(\mathbf{76.0 \pm 0.6}\) & \(\mathbf{69.9 \pm 0.6}\) & \(\mathbf{67.8 \pm 0.5}\) & \(70.1 \pm 0.7\) & \(\mathbf{69.3 \pm 0.1}\)\\
\bottomrule
\end{tabular}
\end{center}

\end{table*}

Power-SMC improves the all-data average over base sampling on all four
backbones and over low-temperature sampling on three of four. \method further achieves the best training-free average
on every backbone, outperforming Power-SMC on 18 of the 20 benchmark results. Without post-training, \method surpasses Game-RL across
all reported benchmarks and remains competitive with TRACE-RL. More specifically, \method performs better on perception-focused benchmarks, while
TRACE-RL performs better on reasoning benchmarks.

\subsection{Component ablation}
All ablations use Qwen2.5-VL-7B-Instruct on five benchmarks. Relative to
\method, w/o islands uses $K=1,M=32$, w/o visual scouts disables scouting, and
w/o answer power sets $\gamma=1$. Additional scout ablations appear in
\cref{app:additional_visual_ablations}. All accuracy analyses use the same
seeded categorical readout as the main results.

\begin{table*}[!htbp]
\caption{Pass@1 (\%) for component ablations at a matched 32-particle budget. All-data averages are
question-weighted.}
\label{tab:component_ablation}
\begin{center}
\scriptsize
\setlength{\tabcolsep}{4pt}
\begin{tabular}{lccccc|c}
\toprule
Variant & LogicVista & MathVista & MMStar-R & MMStar-P & RealWorldQA & All-data avg.\\
\midrule
\method & \(\mathbf{43.53 \pm 2.09}\) & \(\mathbf{71.70 \pm 0.77}\) & \(\mathbf{65.25 \pm 1.04}\) & \(\mathbf{63.05 \pm 0.41}\) & \(\mathbf{70.00 \pm 0.74}\) & \(\mathbf{65.05 \pm 0.33}\)\\
w/o islands & \(40.40 \pm 1.06\) & \(70.68 \pm 0.73\) & \(64.10 \pm 1.76\) & \(62.55 \pm 0.82\) & \(69.97 \pm 0.67\) & \(64.01 \pm 0.63\)\\
w/o visual scouts & \(41.57 \pm 1.04\) & \(71.10 \pm 0.81\) & \(64.65 \pm 0.51\) & \(62.55 \pm 0.72\) & \(\mathbf{70.00 \pm 0.74}\) & \(64.42 \pm 0.26\)\\
w/o answer power & \(43.02 \pm 1.46\) & \(71.13 \pm 1.19\) & \(65.13 \pm 0.99\) & \(62.05 \pm 1.34\) & \(68.04 \pm 0.51\) & \(64.26 \pm 0.33\)\\
\bottomrule
\end{tabular}
\end{center}

\end{table*}

All three components improve the all-data average in \cref{tab:component_ablation}. Island isolation improves every benchmark and provides the largest aggregate gain. Answer-marginal power also improves all five benchmarks. At the default scout fraction, visual scouting improves all four benchmarks on which it activates. RealWorldQA is unchanged because no evaluated run reaches the scout phase (see \cref{tab:generation_horizon} in \cref{app:generation_horizon}).

%% file: sec/6_discussion.tex
\section{Discussion}
\label{sec:discussion}

\subsection{From island coverage to pass@1}
\label{sec:discussion_islands}

We report pass@1 for each system and readout, pass@4 across four independently seeded executions, and Coverage@32, the fraction of terminal populations containing a correct answer. On tasks that reach an intervention checkpoint, the first stage generally raises Coverage@32 and pass@4. The second stage further raises pass@1 by concentrating answer mass, reducing pass@4 on four of five benchmarks as probability mass becomes more concentrated on the leading answers. Cross-benchmark and generation-horizon analyses are given in \cref{app:coverage_all_benchmarks,app:generation_horizon}.

\paragraph{Reasoning.} Reasoning traces remain active beyond the first resampling checkpoint, allowing preserved islands to develop different continuations. Across all three reasoning benchmarks, the first stage raises Coverage@32 and pass@4, while answer-level sharpening further improves pass@1 (\cref{tab:horizon_coverage,app:coverage_all_benchmarks}). The genealogy statistics in \cref{app:island_genealogy} associate this gain with more surviving roots and less mass concentrated in the largest ancestral family.

\paragraph{Perception.}
RealWorldQA is the limiting no-intervention case: every response finishes
before the first checkpoint, and no run activates visual scouting
(\cref{tab:generation_horizon}). On RealWorldQA, the first stages of \method and Power-SMC therefore induce the same output distribution. The small observed differences in
pass@1, pass@4, and Coverage@32 reflect finite-sample variation rather than a
first-stage mechanism. The resulting pass@1 gain mainly comes from the second-stage readout (\cref{tab:horizon_coverage}).

\begin{table*}[!htbp]
\caption{Representative reasoning and perception behavior with
Qwen2.5-VL-7B-Instruct. The $\gamma=1$ rows replay the \method populations
without answer-level sharpening.}
\label{tab:horizon_coverage}
\begin{center}
\scriptsize
\setlength{\tabcolsep}{4.5pt}
\begin{tabular}{llccc}
\toprule
Task & System/readout & Pass@1 & Pass@4 & Coverage@32\\
\midrule
LogicVista & Power-SMC & $42.8\!\pm\!1.0$ & $65.2$ & $63.1\!\pm\!0.8$\\
& \method, $\gamma=1$ & $43.02\!\pm\!1.46$ & $66.3$ & $\mathbf{77.4\!\pm\!1.4}$\\
& \method, $\gamma=2$ & $\mathbf{43.53\!\pm\!2.09}$ & $\mathbf{67.9}$ & $\mathbf{77.4\!\pm\!1.4}$\\
\midrule
RealWorldQA & Power-SMC & $69.2\!\pm\!0.4$ & $81.2$ & $96.5\!\pm\!0.2$\\
& \method, $\gamma=1$ & $68.04\!\pm\!0.51$ & $\mathbf{81.7}$ & $\mathbf{96.6\!\pm\!0.2}$\\
& \method, $\gamma=2$ & $\mathbf{70.00\!\pm\!0.74}$ & $78.3$ & $\mathbf{96.6\!\pm\!0.2}$\\
\bottomrule
\end{tabular}
\end{center}
\end{table*}

\subsection{Trajectory power and answer-level support}
\label{sec:discussion_power}

The two exponents act at different levels. Increasing $\alpha$ favors
individually likely trajectories, whereas applying $\gamma$ after answer
aggregation rewards their collective terminal mass. Our answer-marginal second
stage therefore mediates between a few high-weight trajectories and broader
support from many medium- or low-weight trajectories.
\Cref{tab:readout_pass1} compares answer-marginal power with one-stage sequence power at
$\alpha=2$ and $\alpha=4$.

\begin{table*}[!htbp]
\caption{Pass@1 (\%) under alternative sharpening rules. The all-data average
is question-weighted. Rows sharing $\alpha$ replay the same token-40 populations;
changing $\alpha$ requires separate sampling.}
\label{tab:readout_pass1}
\begin{center}
\scriptsize
\setlength{\tabcolsep}{3.5pt}
\begin{tabular}{lccccc|c}
\toprule
$(\alpha,\gamma)$ & LogicVista & MathVista & MMStar-R & MMStar-P & RealWorldQA & All-data avg.\\
\midrule
$(2,1)$ & $43.02\!\pm\!1.46$ & $71.13\!\pm\!1.19$ & $65.13\!\pm\!0.99$ & $62.05\!\pm\!1.34$ & $68.04\!\pm\!0.51$ & $64.26\!\pm\!0.33$\\
$(4,1)$ & $42.58\!\pm\!1.20$ & $71.25\!\pm\!0.66$ & $64.08\!\pm\!1.05$ & $\mathbf{63.10\!\pm\!0.89}$ & $\mathbf{70.59\!\pm\!0.56}$ & $64.62\!\pm\!0.35$\\
$(2,2)$ & $\mathbf{43.53\!\pm\!2.09}$ & $\mathbf{71.70\!\pm\!0.77}$ & $\mathbf{65.25\!\pm\!1.04}$ & $63.05\!\pm\!0.41$ & $70.00\!\pm\!0.74$ & $\mathbf{65.05\!\pm\!0.33}$\\
\bottomrule
\end{tabular}
\end{center}

\end{table*}

Relative to one-stage $\alpha=2$, answer-marginal second power improves
pass@1 on all five benchmarks. It also outperforms one-stage $\alpha=4$ on
all reasoning benchmarks. As shown in \cref{app:split_sharpening}, squaring the aggregated answer mass
introduces cross terms between trajectories supporting the same answer,
allowing their evidence to reinforce one another. On MMStar-P and RealWorldQA, $(2,2)$ achieves slightly lower pass@1 than one-stage $(4,1)$. \Cref{app:direct_answer_finite_particles} analyzes this behavior in a simplified direct-answer setting.

\subsection{Efficiency}
The principal additional cost is bounded decoding on the temporary scout branches. Answer aggregation requires no additional model forward pass. On a single NVIDIA RTX 5090 GPU with a common Transformers runtime, per-example latency and peak VRAM are $4.11$ seconds and $15.61$ GB for base sampling, $7.85$ seconds and $17.27$ GB for Power-SMC, and $10.37$ seconds and $17.35$ GB for \method. The complete cost table and asymptotic analysis are provided in \cref{app:complexity}.

\subsection{Limitations}
The answer readout cannot repair a wrong or missing first-stage answer mode and
depends on a task-appropriate canonicalizer. For unconstrained free-form
responses, semantically equivalent strings that cannot be merged
deterministically are treated as different answers, fragmenting their marginal
mass. Merging them requires an external semantic-equivalence model or LLM
judge. Without one, distinct strings remain separate. When every trajectory
produces a unique string, answer-level power reduces to particle-wise
trajectory sharpening rather than a separate aggregation stage. Access to
model logits and decoder states restricts the method to open-weight LVLMs.

%% file: sec/7_conclusion.tex
\section{Conclusion}
\label{sec:conclusion}

We first transfer Power-SMC to open-ended LVLM decoding by formulating a
sequence-power target over responses conditioned on the image and prompt,
establishing a strong training-free baseline for power sampling in LVLMs. We then introduce \method to improve finite-particle exploration and
answer-marginal aggregation and sharpening. Island-local resampling preserves
trajectory families that seed prefix-conditioned visual scouts, whose routed
proposals are corrected to the base LVLM target. The second stage aggregates
terminal mass by answer before a finite answer-level power and stochastic draw.
Overall, \method delivers aggregate improvements over Power-SMC on both
reasoning and perception tasks and remains competitive with post-trained
systems, without parameter updates or a learned verifier.

%% file: sec/A_analysis.tex
\section{Additional analysis}
\label{app:additional_analysis}

\subsection{Island genealogy}
\label{app:island_genealogy}
We report terminal-ancestry statistics to characterize the genealogical effect
of island-local resampling. Let $\rho_r$ denote the normalized terminal mass carried by
descendants of initial particle $r$. We report the effective number of roots
$N_{\mathrm{root}}^{\mathrm{eff}}=\exp(-\sum_r\rho_r\log\rho_r)$ and the mass of the
largest root $\max_r\rho_r$. These statistics use the first-stage terminal
weights, before answer-level sharpening.

\begin{table*}[!htbp]
\caption{Terminal genealogy on the reasoning benchmarks with
Qwen2.5-VL-7B-Instruct. The two rows
for each dataset share the same 32-particle budget, visual proposal, and power
parameters.}
\label{tab:island_genealogy}
\begin{center}
\scriptsize
\setlength{\tabcolsep}{5pt}
\begin{tabular}{llcc}
\toprule
Dataset & Population & Effective roots $\uparrow$ & Largest-root mass $\downarrow$\\
\midrule
LogicVista & $1\times32$ (w/o islands) & $1.625\!\pm\!0.048$ & $0.900\!\pm\!0.006$\\
& $4\times8$ (\method) & $\mathbf{1.859\!\pm\!0.051}$ & $\mathbf{0.844\!\pm\!0.010}$\\
\midrule
MathVista & $1\times32$ (w/o islands) & $3.790\!\pm\!0.030$ & $0.671\!\pm\!0.004$\\
& $4\times8$ (\method) & $\mathbf{4.005\!\pm\!0.050}$ & $\mathbf{0.648\!\pm\!0.005}$\\
\midrule
MMStar-R & $1\times32$ (w/o islands) & $2.628\!\pm\!0.027$ & $0.789\!\pm\!0.004$\\
& $4\times8$ (\method) & $\mathbf{2.832\!\pm\!0.033}$ & $\mathbf{0.754\!\pm\!0.008}$\\
\bottomrule
\end{tabular}
\end{center}

\end{table*}

Across all three reasoning benchmarks, island isolation increases the effective
number of terminal roots and reduces the largest ancestral family's mass. The
effect is obtained at the same total particle budget and with the same visual
and power parameters, so it directly measures the consequence of replacing
global resampling with island-local resampling. The result supports the intended
mechanism: a high-weight family can expand within its island without erasing
promising families maintained by the other islands.

\subsection{Coverage, pass@1, and pass@4 across benchmarks}
\label{app:coverage_all_benchmarks}
\Cref{tab:coverage_all_benchmarks} extends the comparison in
\cref{tab:horizon_coverage} across the remaining benchmarks.

\begin{table*}[!htbp]
\caption{Realized pass@1, pass@4, and Coverage@32 (\%) with
Qwen2.5-VL-7B-Instruct. Pass@1 and Coverage@32 are mean $\pm$ sample standard
deviation; pass@4 measures success across four independently seeded complete executions.}
\label{tab:coverage_all_benchmarks}
\begin{center}
\scriptsize
\setlength{\tabcolsep}{4.5pt}
\begin{tabular}{llccc}
\toprule
Benchmark & System & Pass@1 & Pass@4 & Coverage@32\\
\midrule
MathVista & Power-SMC, $1\times32$ & $70.2\!\pm\!0.7$ & $80.4$ & $81.3\!\pm\!0.4$\\
MathVista & \method, $\gamma=1$ & $71.13\!\pm\!1.19$ & $\mathbf{81.9}$ & $\mathbf{86.6\!\pm\!0.3}$\\
MathVista & \method, $\gamma=2$ & $\mathbf{71.70\!\pm\!0.77}$ & $81.3$ & $\mathbf{86.6\!\pm\!0.3}$\\
\midrule
MMStar-R & Power-SMC, $1\times32$ & $63.7\!\pm\!0.8$ & $78.4$ & $76.3\!\pm\!0.9$\\
MMStar-R & \method, $\gamma=1$ & $65.13\!\pm\!0.99$ & $\mathbf{78.9}$ & $\mathbf{85.5\!\pm\!1.0}$\\
MMStar-R & \method, $\gamma=2$ & $\mathbf{65.25\!\pm\!1.04}$ & $78.2$ & $\mathbf{85.5\!\pm\!1.0}$\\
\midrule
MMStar-P & Power-SMC, $1\times32$ & $61.7\!\pm\!0.9$ & $\mathbf{72.2}$ & $89.3\!\pm\!1.0$\\
MMStar-P & \method, $\gamma=1$ & $62.05\!\pm\!1.34$ & $72.0$ & $\mathbf{90.0\!\pm\!0.4}$\\
MMStar-P & \method, $\gamma=2$ & $\mathbf{63.05\!\pm\!0.41}$ & $68.8$ & $\mathbf{90.0\!\pm\!0.4}$\\
\bottomrule
\end{tabular}
\end{center}

\end{table*}

Together, these tables show that, before answer-level sharpening, \method attains higher empirical pass@4 than
Power-SMC on four of five benchmarks while also increasing Coverage@32. Applying $\gamma=2$ subsequently improves pass@1 over the
$\gamma=1$ readout on every benchmark, but lowers pass@4 on four of the five benchmarks as probability mass becomes more concentrated on the leading answers. LogicVista is the exception, improving both pass@1 and pass@4.

\subsection{Generation horizon and scout activation}
\label{app:generation_horizon}
The fraction active at a checkpoint counts particles that have not emitted EOS.
Completed particles continue through the exponent bridge and checkpoint
calculation, but resampling them cannot create new continuations. As shown in
\cref{tab:generation_horizon}, nearly every reasoning particle remains active
through both checkpoints, whereas direct-answer responses are usually complete
before resampling and visual scouting.

\begin{table*}[!htbp]
\caption{Generation activity and scout activation. ``Scout
questions'' is the fraction of question--seed runs in which at least one
particle activates visual scouting.}
\label{tab:generation_horizon}
\begin{center}
\scriptsize
\setlength{\tabcolsep}{5pt}
\begin{tabular}{lccc}
\toprule
Benchmark & Active at 32 (\%) & Active at 40 (\%) & Scout questions (\%)\\
\midrule
LogicVista & 100.0 & 100.0 & 100.0\\
MathVista & 100.0 & 99.9 & 100.0\\
MMStar-R & 100.0 & 100.0 & 100.0\\
MMStar-P & 5.9 & 4.7 & 31.2\\
RealWorldQA & 0.0 & 0.0 & 0.0\\
\bottomrule
\end{tabular}
\end{center}

\end{table*}




\subsection{Additional visual-stage ablations}
\label{app:additional_visual_ablations}
\Cref{tab:additional_visual_ablations} tests the region-specific attention
bias and the scope of the overlap penalty. The global-attention-only variant
removes the region-specific second bias. The island-local penalty variant
changes only the scope of the overlap penalty, retaining the default
coefficient $\mu=1$ and all other settings.

\begin{table*}[!htbp]
\caption{Pass@1 (\%) for additional visual-stage ablations with
Qwen2.5-VL-7B-Instruct.}
\label{tab:additional_visual_ablations}
\begin{center}
\scriptsize
\setlength{\tabcolsep}{3.5pt}
\begin{tabular}{lccccc|c}
\toprule
Variant & LogicVista & MathVista & MMStar-R & MMStar-P & RealWorldQA & All-data avg.\\
\midrule
\method & \(43.53 \pm 2.09\) & \(71.70 \pm 0.77\) & \(65.25 \pm 1.04\) & \(63.05 \pm 0.41\) & \(70.00 \pm 0.74\) & \(65.05 \pm 0.33\)\\
global attention only & \(42.69 \pm 1.28\) & \(70.75 \pm 0.76\) & \(64.65 \pm 0.56\) & \(62.65 \pm 0.84\) & \(70.00 \pm 0.74\) & \(64.48 \pm 0.19\)\\
island-local overlap penalty & \(42.80 \pm 1.36\) & \(71.58 \pm 1.35\) & \(65.00 \pm 1.28\) & \(63.20 \pm 0.59\) & \(70.00 \pm 0.74\) & \(64.88 \pm 0.69\)\\

\bottomrule
\end{tabular}
\end{center}

\end{table*}

The region-specific bias improves pass@1 over global attention only on every
benchmark where scouts activate. With $\mu=1$, global overlap penalization
improves the all-data average over island-local penalization from $64.88\%$ to
$65.05\%$ and gives higher means on three of the four scout-active benchmarks.
RealWorldQA is unchanged because no scout activates there.

\paragraph{Overlap-penalty coefficient.}
\Cref{tab:overlap_penalty_sensitivity} varies the global overlap coefficient
$\mu$ while retaining the biased route and all other settings. Among the
evaluated values, $\mu=1$ gives the highest
question-weighted average; $\mu=0.5$ gives the highest MathVista and MMStar-P
means.

\begin{table*}[!htbp]
\caption{Global overlap-penalty sensitivity in realized pass@1 (\%) with
Qwen2.5-VL-7B-Instruct. Values are mean $\pm$ sample standard deviation over
four seeds; the all-data average is question-weighted.}
\label{tab:overlap_penalty_sensitivity}
\begin{center}
\scriptsize
\setlength{\tabcolsep}{4pt}
\begin{tabular}{lccccc|c}
\toprule
Overlap coefficient & LogicVista & MathVista & MMStar-R & MMStar-P & RealWorldQA & All-data avg.\\
\midrule
$\mu=0$ & $43.14\!\pm\!0.71$ & $71.38\!\pm\!1.76$ & $65.03\!\pm\!1.35$ & $62.65\!\pm\!0.77$ & $70.00\!\pm\!0.74$ & $64.80\!\pm\!0.66$\\
$\mu=0.5$ & $43.02\!\pm\!1.79$ & $\mathbf{71.80\!\pm\!0.62}$ & $64.45\!\pm\!0.29$ & $\mathbf{63.25\!\pm\!0.53}$ & $70.00\!\pm\!0.74$ & $64.83\!\pm\!0.37$\\
$\mu=1$ & $\mathbf{43.53\!\pm\!2.09}$ & $71.70\!\pm\!0.77$ & $\mathbf{65.25\!\pm\!1.04}$ & $63.05\!\pm\!0.41$ & $70.00\!\pm\!0.74$ & $\mathbf{65.05\!\pm\!0.33}$\\
\bottomrule
\end{tabular}
\end{center}
\end{table*}

\paragraph{Region-size normalization.}
\Cref{tab:area_exponent_sensitivity} compares the area exponent $\zeta$,
which sets the balance between total attention mass and average per-token
relevance in region scoring. The intermediate value
$\zeta=0.75$ gives the highest question-weighted average and is used by
default.

\begin{table*}[!htbp]
\caption{Area-exponent sensitivity in realized pass@1 (\%).}
\label{tab:area_exponent_sensitivity}
\begin{center}
\scriptsize
\setlength{\tabcolsep}{4pt}
\begin{tabular}{lccccc|c}
\toprule
Area exponent & LogicVista & MathVista & MMStar-R & MMStar-P & RealWorldQA & All-data avg.\\
\midrule
$\zeta=0.5$ & $42.47\!\pm\!1.76$ & $70.85\!\pm\!0.87$ & $64.60\!\pm\!1.04$ & $62.60\!\pm\!0.85$ & $70.00\!\pm\!0.74$ & $64.46\!\pm\!0.65$\\
$\zeta=0.75$ & $\mathbf{43.53\!\pm\!2.09}$ & $\mathbf{71.70\!\pm\!0.77}$ & $\mathbf{65.25\!\pm\!1.04}$ & $\mathbf{63.05\!\pm\!0.41}$ & $70.00\!\pm\!0.74$ & $\mathbf{65.05\!\pm\!0.33}$\\
$\zeta=1$ & $42.13\!\pm\!1.76$ & $70.48\!\pm\!0.58$ & $64.63\!\pm\!0.90$ & $62.75\!\pm\!0.75$ & $70.00\!\pm\!0.74$ & $64.34\!\pm\!0.38$\\
\bottomrule
\end{tabular}
\end{center}

\end{table*}

\paragraph{Scout fraction.}
\Cref{tab:scout_fraction_sensitivity} varies the per-island scout fraction
while keeping the particle budget and all other visual settings fixed. The
default $\rho_V=0.25$ gives the strongest
LogicVista, MathVista, and MMStar-R results and the highest all-data average.
At $\rho_V=0.5$, these three scores decline slightly
but remain above the visual-off variant.

\begin{table*}[!htbp]
\caption{Scout-fraction sensitivity in realized pass@1 (\%).}
\label{tab:scout_fraction_sensitivity}
\begin{center}
\scriptsize
\setlength{\tabcolsep}{4pt}
\begin{tabular}{lccccc|c}
\toprule
Scout fraction & LogicVista & MathVista & MMStar-R & MMStar-P & RealWorldQA & All-data avg.\\
\midrule
visual-off & $41.57\!\pm\!1.04$ & $71.10\!\pm\!0.81$ & $64.65\!\pm\!0.51$ & $62.55\!\pm\!0.72$ & $70.00\!\pm\!0.74$ & $64.42\!\pm\!0.26$\\
$\rho_V=0.125$ & $43.08\!\pm\!0.87$ & $71.18\!\pm\!0.15$ & $64.13\!\pm\!0.75$ & $\mathbf{63.05\!\pm\!1.22}$ & $70.00\!\pm\!0.74$ & $64.55\!\pm\!0.21$\\
$\rho_V=0.25$ & $\mathbf{43.53\!\pm\!2.09}$ & $\mathbf{71.70\!\pm\!0.77}$ & $\mathbf{65.25\!\pm\!1.04}$ & $\mathbf{63.05\!\pm\!0.41}$ & $70.00\!\pm\!0.74$ & $\mathbf{65.05\!\pm\!0.33}$\\
$\rho_V=0.5$ & $43.36\!\pm\!0.80$ & $71.28\!\pm\!0.57$ & $64.80\!\pm\!0.55$ & $62.10\!\pm\!1.09$ & $70.00\!\pm\!0.74$ & $64.66\!\pm\!0.42$\\
\bottomrule
\end{tabular}
\end{center}

\end{table*}

\FloatBarrier

%% file: sec/B_appendix.tex
\section{Importance-weight derivation}
\label{app:weights}
Suppressing the fixed conditioning pair $(I,x)$, let
$p_{F,t}(y_{1:t})$ denote the base-model likelihood accumulated by step $t$,
with the likelihood held fixed after a trajectory ends. For
$\varphi_t=p_{F,t}^{\beta_t}$, the generic SMC increment, with proposal-history dependence suppressed, is
\begin{equation}
G_t=\frac{\varphi_t(y_{1:t})}
{\varphi_{t-1}(y_{<t})q_t(y_t\mid y_{<t})}.
\end{equation}

\paragraph{Meaning of the proposal.}
Let $\mathcal F_{t-1}$ denote the information available to the sampler
immediately before token $t$ is drawn. It contains all particle prefixes,
weights, island masses, resampling ancestry, completion states, and any scout
routes and schedules selected by that time. Thus, $\mathcal F_{t-1}$ is an
information state maintained by the SMC controller.
For particle $(k,m)$, the model is evaluated with the
standard autoregressive context $(I,x,y_{<t}^{k,m})$, represented
computationally by its decoder KV cache and attention mask. Population-level
information in $\mathcal F_{t-1}$ is used outside the LVLM to select the
proposal branch and, for a scout, its routed attention bias. We display
$y_{<t}^{k,m}$ separately in the conditioning notation to emphasize
autoregressive dependence, although it is contained in $\mathcal F_{t-1}$.

Let $\mathcal S_{t-1}$ be the particles whose scout window is active before
step $t$. The next token is drawn from the realized proposal
\begin{equation}
q_t^{k,m}(v\mid y_{<t}^{k,m},\mathcal F_{t-1})
=
\begin{cases}
q_t^{A,k,m}(v\mid y_{<t}^{k,m},I,x),
& (k,m)\in\mathcal S_{t-1},\\[3pt]
q_t^F(v\mid y_{<t}^{k,m},I,x),
& \text{otherwise}.
\end{cases}
\label{eq:realized_particle_proposal}
\end{equation}
We write $q_t^F$ when the particle indices are clear from the conditioned
prefix. It is the powered native next-token conditional, whereas
$q_t^{A,k,m}$ is the powered conditional produced by the routed attention
bias. See \cref{eq:local_power_proposal,eq:attention_proposal}. An active scout
has $q_t^{A,k,m}$ as its next-token proposal. The realized-token probability appears in the denominator of $G_t$, ensuring that
a continuation does not receive excess target mass merely because its proposal
samples it more often.

\paragraph{From a path weight to an incremental weight.}
For a fixed ancestral path, suppress the particle indices and let
\begin{equation}
Q_t(y_{1:t})
=Q_{t-1}(y_{<t})q_t(y_t\mid y_{<t},\mathcal F_{t-1}),
\qquad Q_0=1,
\label{eq:joint_proposal_factorization}
\end{equation}
be its joint proposal probability. Iterating this recursion gives
$Q_t(y_{1:t})=\prod_{s=1}^t q_s(y_s\mid y_{<s},\mathcal F_{s-1})$.
Let $\pi_t(y_{1:t})=\varphi_t(y_{1:t})/Z_t$ denote the normalized target,
where $Z_t=\sum_{y_{1:t}}\varphi_t(y_{1:t})$. Under standard importance
sampling~\citep{doucet2011tutorial}, a prefix drawn from $Q_t$ receives weight
\begin{equation}
\frac{\pi_t(y_{1:t})}{Q_t(y_{1:t})}
=\frac{\varphi_t(y_{1:t})}{Z_tQ_t(y_{1:t})}.
\label{eq:normalized_path_weight}
\end{equation}
The normalizer $Z_t$ is generally unavailable, but it is common to every
particle and therefore cancels when the weights are normalized. Ignoring this
common factor and the initial particle weight $1/M$,
the cumulative unnormalized importance weight is
\begin{equation}
\mathcal W_t(y_{1:t})
:=\frac{\varphi_t(y_{1:t})}{Q_t(y_{1:t})}.
\label{eq:path_importance_weight}
\end{equation}

This ratio has a direct interpretation. A prefix is sampled with probability
$Q_t(y_{1:t})$ and receives weight
$\mathcal W_t(y_{1:t})=\varphi_t(y_{1:t})/Q_t(y_{1:t})$. Multiplying sampling
probability by weight gives
$Q_t(y_{1:t})\mathcal W_t(y_{1:t})=\varphi_t(y_{1:t})$: the proposal probability
cancels path by path, leaving exactly the unnormalized target mass.

Here $\mathcal W_t$ is only an auxiliary cumulative weight used to derive
$G_t$. The implemented particle weights $w_t^{k,m}$ follow the same incremental
update and are reset after resampling. Hence,
\begin{equation}
\begin{aligned}
\frac{\mathcal W_t(y_{1:t})}{\mathcal W_{t-1}(y_{<t})}
&=\frac{\varphi_t(y_{1:t})}{\varphi_{t-1}(y_{<t})}
  \frac{Q_{t-1}(y_{<t})}{Q_t(y_{1:t})}\\
&=\frac{\varphi_t(y_{1:t})}
{\varphi_{t-1}(y_{<t})q_t(y_t\mid y_{<t},\mathcal F_{t-1})}=G_t.
\end{aligned}
\label{eq:incremental_weight_derivation}
\end{equation}

At the one-step level, the same cancellation is immediate. A token $v$ is
sampled with probability $q_t(v\mid y_{<t},\mathcal F_{t-1})$, while its
incremental weight contains the reciprocal of this probability. Their product
therefore cancels $q_t$ token by token. This correction is valid for any
proposal chosen from the available population history, provided that it assigns
positive probability wherever the target increment is positive.

Before the response ends, the intermediate targets satisfy
\[
\varphi_t(y_{1:t})
=
p_F(y_{1:t}\mid I,x)^{\beta_t},
\qquad
\varphi_{t-1}(y_{<t})
=
p_F(y_{<t}\mid I,x)^{\beta_{t-1}}.
\]
Using the autoregressive factorization
\[
p_F(y_{1:t}\mid I,x)
=
p_F(y_{<t}\mid I,x)
p_F(y_t\mid y_{<t},I,x)
\]
in the incremental-weight definition yields
\begin{equation}
G_t
=
\frac{
p_F(y_t\mid y_{<t},I,x)^{\beta_t}
p_F(y_{<t}\mid I,x)^{\beta_t-\beta_{t-1}}
}{
q_t(y_t\mid y_{<t})
}.
\label{eq:appendix_increment}
\end{equation}

\paragraph{The two corrections under the base proposal.}
Equation~\eqref{eq:appendix_increment} follows by writing
$p_F(y_{1:t}\mid I,x)=p_F(y_{<t}\mid I,x)
 p_F(y_t\mid y_{<t},I,x)$ and dividing the target at exponent $\beta_t$
by the preceding target at exponent $\beta_{t-1}$. To see what the importance
weight corrects, define the prefix-dependent local normalizer
\begin{equation}
\begin{aligned}
Z_t^{\mathrm{loc}}(y_{<t})
&:=\sum_{u\in\mathcal V}p_F(u\mid y_{<t},I,x)^{\beta_t},\\
q_t^F(y_t\mid y_{<t},I,x)
&=\frac{p_F(y_t\mid y_{<t},I,x)^{\beta_t}}
{Z_t^{\mathrm{loc}}(y_{<t})}.
\end{aligned}
\label{eq:appendix_local_proposal}
\end{equation}
Substituting $q_t=q_t^F$ into
\cref{eq:appendix_increment} cancels the current-token power and gives
\begin{equation}
G_t
=Z_t^{\mathrm{loc}}(y_{<t})
 p_F(y_{<t}\mid I,x)^{\beta_t-\beta_{t-1}}.
\label{eq:appendix_full_increment}
\end{equation}
The two remaining factors have distinct roles. First, independently sampling
each token from the locally normalized $q_t^F$ would place the
prefix-dependent factor $1/Z_t^{\mathrm{loc}}(y_{<t})$ in the joint proposal.
Multiplying the particle weight by $Z_t^{\mathrm{loc}}(y_{<t})$ cancels this
factor. Otherwise, local temperature sampling would generally define a
different distribution from the desired power of the complete sequence.
Second, the prefix was evaluated under exponent $\beta_{t-1}$ at the preceding
step, whereas the new intermediate target evaluates the entire extended prefix
under $\beta_t$. The factor
$p_F(y_{<t}\mid I,x)^{\beta_t-\beta_{t-1}}$ supplies exactly this missing
exponent difference to all previously generated tokens.

For an active scout, $y_t^{k,m}$ is sampled from
$q_t^{A,k,m}(\cdot\mid y_{<t}^{k,m})$, while $\varphi_t$ remains defined by the
base LVLM. Along the particle's ancestral path, the joint proposal
probability therefore extends as
\[
Q_t(y_{1:t}^{k,m})
=Q_{t-1}(y_{<t}^{k,m})
q_t^{A,k,m}(y_t^{k,m}\mid y_{<t}^{k,m}).
\]
Substituting this factorization into the ratio of consecutive cumulative
importance weights gives
\begin{equation}
\begin{aligned}
G_t^{A,k,m}
&=\frac{\mathcal W_t(y_{1:t}^{k,m})}
{\mathcal W_{t-1}(y_{<t}^{k,m})}\\
&=\frac{\varphi_t(y_{1:t}^{k,m})/Q_t(y_{1:t}^{k,m})}
{\varphi_{t-1}(y_{<t}^{k,m})/Q_{t-1}(y_{<t}^{k,m})}\\
&=\frac{\varphi_t(y_{1:t}^{k,m})}
{\varphi_{t-1}(y_{<t}^{k,m})
q_t^{A,k,m}(y_t^{k,m}\mid y_{<t}^{k,m})}.
\end{aligned}
\label{eq:scout_proposal_correction}
\end{equation}
The last equality cancels the preceding-path proposal probability
$Q_{t-1}$. Consequently,
\[
\varphi_{t-1}(y_{<t}^{k,m})
q_t^{A,k,m}(y_t^{k,m}\mid y_{<t}^{k,m})G_t^{A,k,m}
=\varphi_t(y_{1:t}^{k,m}).
\]
Thus attention reactivation changes how often each continuation is proposed,
while importance weighting maps its mass back to the same base LVLM
intermediate target.

\paragraph{Completed trajectories.}
Suppose a response emits EOS at step $e<t$. The completed sequence $y_{1:e}$
then remains fixed: no new token is drawn, so there is no new proposal
probability to divide out. The update therefore reduces
to
\begin{equation}
\begin{aligned}
G_t
&=\frac{\varphi_t(y_{1:e})}{\varphi_{t-1}(y_{1:e})}\\
&=p_F(y_{1:e}\mid I,x)^{\beta_t-\beta_{t-1}}.
\end{aligned}
\end{equation}

Multiplying these post-EOS increments from $e+1$ through $H$ telescopes the
completed sequence's exponent from $\beta_e$ to $\beta_H=\alpha$. Thus every
terminal response is ultimately weighted by its base-model likelihood raised
to the same exponent $\alpha$, regardless of when it emitted EOS. Response length affects the model likelihood, but not the terminal target exponent.

\subsection{Island resampling and terminal mass}
\label{app:island_mechanics}
For completeness, the normalized base proposal used outside the scout
episode is
\begin{equation}
q_t^F(v\mid y_{<t},I,x)
=\frac{p_F(v\mid y_{<t},I,x)^{\beta_t}}
{\sum_u p_F(u\mid y_{<t},I,x)^{\beta_t}}.
\label{eq:local_power_proposal}
\end{equation}
Within island $k$, weights are normalized as
$\bar w_t^{k,m}=w_t^{k,m}/\sum_{j=1}^M w_t^{k,j}$. At a checkpoint, the
island resamples if
\begin{equation}
\ESS_{k,t}
=\left(\sum_{m=1}^M(\bar w_t^{k,m})^2\right)^{-1}
<\rho_wM,
\label{eq:weightess}
\end{equation}
where $\rho_w\in(0,1]$ is the threshold fraction. Stratified
resampling law draws $M$ parents from $\bar w_t^{k,1:M}$ and resets their
weights to $1/M$. If the test is not triggered, the particles and their
weights pass through unchanged. In either case, ancestry is confined to the
island:
\begin{equation}
\operatorname{Anc}(k,m,t)
\subseteq\{k\}\times\{1,\ldots,M\}.
\label{eq:island_ancestry}
\end{equation}
Confining ancestry within each island preserves $K$ separate genealogies.

Each island maintains the standard SMC normalizing-constant estimate. With the
pre-update normalized weights, define
\begin{align}
c_{k,t}
&=\sum_{m=1}^M\bar w_{t-1}^{k,m}G_t^{k,m},\\
\widehat Z_{k,t}
&=\widehat Z_{k,t-1}c_{k,t},
\qquad \widehat Z_{k,0}=1.
\label{eq:island_mass_eval}
\end{align}
At the terminal bridge exponent, the normalized island mass is
\begin{equation}
\Omega_{k,H}
=\frac{\widehat Z_{k,H}}
{\sum_{j=1}^K\widehat Z_{j,H}}.
\label{eq:island_probability}
\end{equation}
An ordinary hierarchical Island-SMC output first draws an island and then a
particle within it:
\begin{align}
k^\star
&\sim\operatorname{Cat}(\Omega_{1:K,H}),
\label{eq:island_draw}\\
m^\star\mid k^\star
&\sim\operatorname{Cat}(\bar w_H^{k^\star,1:M}).
\label{eq:particle_draw}
\end{align}
Multiplying these two probabilities gives the pooled terminal mass in
\cref{eq:island_terminal_weight}. Thus the normalizer estimates compare the
target mass represented by different islands without allowing cross-island
resampling.

\section{Visual-scout routing specification}
\label{app:routing_spec}
The visual intervention occurs after token $\tau_{\mathrm{vis}}$ has been
weighted and before the next response token is sampled. We permit
$\tau_{\mathrm{vis}}\geq1$. The implementation requires a generated token whose
query can condition the route. The $L_{\mathrm{vis}}$ scout transitions finish before the next
island-local resampling checkpoint. If no island contains at least two
unfinished particles, routing is skipped.

Let $\mathcal U_k$ be the unfinished particles in island $k$ at the
checkpoint. Its scout quota is
\begin{equation}
B_k=\min\left\{
\left\lceil\rho_VM\right\rceil,
\max(|\mathcal U_k|-1,0)
\right\},
\qquad B=\sum_{k=1}^K B_k.
\label{eq:scout_budget}
\end{equation}
Hence an island with enough unfinished particles receives the common quota,
whereas a smaller active island retains at least one base-proposal anchor. The
quota is independent of island ESS.

For a fixed decoder layer $\ell$, the current text query and original-image
visual keys define
\begin{equation}
\xi_{i,a}^{k,m}
=\frac{\exp(\langle q_a^{\ell,k,m},k_{i,a}^{\ell}\rangle/\sqrt{d_h})}
{\sum_{j\in\mathcal V_I}
\exp(\langle q_a^{\ell,k,m},k_{j,a}^{\ell}\rangle/\sqrt{d_h})}.
\label{eq:visual_qk_proxy}
\end{equation}
Here $\mathcal V_I$ is the visual-token index set,
$a\in\{1,\ldots,N_h\}$ indexes attention heads, and $d_h$ is the per-head
key dimension. The softmax is restricted to visual tokens, so
$\xi_{i,a}^{k,m}$ records where the current response prefix looks within the
image. The score in \cref{eq:particle_region_score} aggregates this visual-only
relevance over heads and the visual tokens covered by each candidate region.
The exponent $\zeta=0$ uses total relevance mass and favors larger regions,
while $\zeta=1$ uses average relevance per visual token. We use $\zeta=0.75$ to partially normalize area, balancing the large-region preference of total relevance against the sensitivity of full averaging to compact high-attention regions.

At the routing checkpoint, the pooled mass of particle $(k,m)$ is
\begin{equation}
\widetilde W_{\tau_{\mathrm{vis}}}^{k,m}
=\frac{\widehat Z_{k,\tau_{\mathrm{vis}}}
\bar w_{\tau_{\mathrm{vis}}}^{k,m}}
{\sum_{j=1}^K\widehat Z_{j,\tau_{\mathrm{vis}}}}.
\label{eq:intermediate_particle_mass}
\end{equation}
This is the checkpoint analogue of
\cref{eq:island_terminal_weight}. After computing
\cref{eq:particle_region_utility}, let $u_{\min}$ and $u_{\max}$ be the
extrema over all initially eligible particle--region pairs. We use
\begin{equation}
\widetilde u_{k,m,g}
=\frac{u_{k,m,g}-u_{\min}}
{u_{\max}-u_{\min}+\epsilon},
\qquad \epsilon=10^{-12},
\label{eq:normalized_pair_utility}
\end{equation}
where the constant only prevents division by zero when every utility is equal.

Let $\mathcal P_{b-1}$ be the particles selected in the first $b-1$ routing
steps, with $\mathcal P_0=\varnothing$, and let $n_k(\mathcal P)$ count
selected particles from island $k$. The candidates for step $b$ are
\begin{equation}
\mathcal C_b=\left\{(k,m,g):
(k,m)\in\mathcal U_k\setminus\mathcal P_{b-1},\quad
n_k(\mathcal P_{b-1})<B_k
\right\}.
\label{eq:routing_candidates}
\end{equation}
This set permits at most one view per particle and at most $B_k$ scouts from
island $k$. Applying \cref{eq:greedy_route} and updating
$\mathcal P_b=\mathcal P_{b-1}\cup\{(k_b,m_b)\}$ produces $B$ assignments.
Maintaining each region's maximum overlap with previously selected regions
gives $O(BKMG)$ greedy selection.



\section{Unbiasedness and exact target correction}
\label{app:routingproof}
Let $\mathcal F_t^-$ denote the sigma-field generated by the complete
algorithmic history through propagation and weighting at step $t$, including
the current particles, normalized weights, and normalizing-constant estimates,
but before the step-$t$ resampling decision. Let $\mathcal F_t$ additionally
include the resulting ancestry and proposal decisions. At
$\tau_{\mathrm{vis}}$, particle selection, region assignment, attention-bias
construction, cache branching, and one-token replay are included in
$\mathcal F_{\tau_{\mathrm{vis}}}$ and affect proposals beginning at
$\tau_{\mathrm{vis}}+1$.

We assume that each proposal is normalized and has support wherever the target
increment is positive, and that token draws are conditionally independent given
the population history. For the analysis, a response that emits EOS at step
$e$ is padded at every later step by a deterministic absorbing EOS transition.
This transition has proposal probability one, leaves the completed response
$y_{1:e}$ unchanged, and uses the bridge increment
$\varphi_t(y_{1:e})/\varphi_{t-1}(y_{1:e})$. All other transitions from that
completed state have zero target mass. Thus the token-step sums below cover
active and completed particles within the same recursion. Write
$\mathcal V_t(y)$ for the available transitions from state $y$: it is the full
vocabulary for an active prefix and the singleton absorbing EOS transition for
a completed response.

The implementation applies stratified resampling separately within each
island. When the ESS rule triggers resampling, the stratified-resampling
procedure constructs the ancestor indices as follows:
\begin{equation}
\begin{aligned}
C_0&=0,\qquad
C_j=\sum_{\ell=1}^j\bar w_t^{k,\ell},\qquad
I_j=[C_{j-1},C_j), && j=1,\ldots,M,\\
S_m&=[(m-1)/M,m/M),\qquad
U_m\overset{\mathrm{ind}}{\sim}\operatorname{Unif}(S_m), && m=1,\ldots,M,\\
A_{k,t}^m&=j\quad\text{when }U_m\in I_j,\\
N_j&=\sum_{m=1}^M\mathbf 1\{A_{k,t}^m=j\}.
\end{aligned}
\label{eq:stratified_construction}
\end{equation}
Here $I_j$ is the cumulative-weight interval assigned to particle $j$,
$S_m$ is the $m$-th equal stratum, and $N_j$ is the number of offspring of
particle $j$. Conditional on $\mathcal F_t^-$,
\begin{equation}
\mathbb E[N_j\mid\mathcal F_t^-]
=\sum_{m=1}^M M\lvert S_m\cap I_j\rvert
=M\lvert I_j\rvert
=M\bar w_t^{k,j},
\label{eq:stratified_expected_offspring}
\end{equation}
where $\lvert\cdot\rvert$ denotes interval length. Consequently, for every
bounded function $f$,
\begin{equation}
\begin{aligned}
&\mathbb E\!\left[
\frac{1}{M}\sum_{m=1}^M
f\!\left(y_{1:t}^{k,A_{k,t}^m}\right)
\middle|\mathcal F_t^-\right]\\
&\qquad=\sum_{j=1}^M\bar w_t^{k,j}f(y_{1:t}^{k,j}).
\end{aligned}
\label{eq:local_resampling_unbiased}
\end{equation}
This proves the conditional unbiasedness of stratified resampling. Because
$A_{k,t}^m\in\{1,\ldots,M\}$ indexes only particles in island $k$, resampling
cannot introduce cross-island ancestors. When resampling is not triggered, the
current particles and normalized weights are carried forward unchanged, so
their weighted empirical measure is preserved deterministically. When $\bar w_t^{k,m}$ subsequently appears as the weight entering step $t+1$, it therefore equals $1/M$ for a resampled descendant and remains the current normalized weight otherwise.

For particle $i=(k,m)$, write its proposal as
$q_t^i(v\mid\mathcal F_{t-1})$. For any bounded function $f$,
\begin{align}
&\mathbb E\!\left[
G_t^i(Y_t^i)f((y_{<t}^i,Y_t^i))
\mid\mathcal F_{t-1}\right]\nonumber\\
&=\sum_{v\in\mathcal V_t(y_{<t}^i)}
\frac{\varphi_t((y_{<t}^i,v))}{\varphi_{t-1}(y_{<t}^i)}
f((y_{<t}^i,v)).
\label{eq:conditional_importance_identity}
\end{align}
The proposal cancels conditionally even though visual routing depends on the
joint population at the preceding checkpoint. Between resampling events, each
particle therefore follows ordinary sequential importance sampling: it extends
its current prefix according to its actual sequence of token proposals and
multiplies its weight by the corresponding incremental importance ratios.
Conditional on the incoming weighted population, the expected updated weighted
measure equals the target update of that empirical measure. Combined with
conditionally unbiased resampling and the island normalizing-mass recursion,
this establishes the unbiasedness of the pooled unnormalized estimator in
\cref{eq:pooled_unbiasedness}.

After propagation and importance-weight updating at step $t$, and before any
optional resampling, define the pooled unnormalized empirical measure
\begin{equation}
\widehat\Gamma_t^{K,M}(f)
=\frac{1}{K}\sum_{k=1}^K\widehat Z_{k,t}
\sum_{m=1}^M\bar w_t^{k,m}f(y_{1:t}^{k,m})
\label{eq:pooled_unnormalized_measure}
\end{equation}
and $\Gamma_t(f)=\sum_{y_{1:t}}\varphi_t(y_{1:t})f(y_{1:t})$.

\paragraph{Proposition 1 (unbiased unnormalized measure).}
For every bounded $f$,
\begin{equation}
\mathbb E\!\left[\widehat\Gamma_t^{K,M}(f)\right]=\Gamma_t(f).
\label{eq:pooled_unbiasedness}
\end{equation}
Consequently, $K^{-1}\sum_{k=1}^K\widehat Z_{k,t}$ is an unbiased estimator of
the normalizing constant $Z_t$.

\begin{proof}
For each island $k$, write
\[
\widehat\Gamma_{k,t}^{M}(f)
:=\widehat Z_{k,t}\sum_{m=1}^M
\bar w_t^{k,m}f(y_{1:t}^{k,m}),
\]
so that
$\widehat\Gamma_t^{K,M}(f)=K^{-1}\sum_{k=1}^K
\widehat\Gamma_{k,t}^{M}(f)$.

\noindent\textbf{Induction statement.}
We prove that, for every island $k$ and every bounded test function $f$ on
the step-$t$ prefix space,
\begin{equation}
\mathbb E\!\left[\widehat\Gamma_{k,t}^{M}(f)\right]
=\Gamma_t(f)
\label{eq:island_unbiasedness}
\end{equation}
by induction on $t$.

\noindent\textbf{Base case ($t=0$).}
Every island contains the empty prefix with normalized weights $1/M$ and
$\widehat Z_{k,0}=1$. Hence
$\widehat\Gamma_{k,0}^{M}(f)=f(\varnothing)=\Gamma_0(f)$.

\noindent\textbf{Inductive step ($t-1$ to $t$).}
Assume \cref{eq:island_unbiasedness} holds at step $t-1$. Before optional
resampling at step $t$, the particle-weight and island-mass recursions give
\begin{align}
\widehat\Gamma_{k,t}^{M}(f)
&=\widehat Z_{k,t-1}c_{k,t}\sum_{m=1}^M
\frac{\bar w_{t-1}^{k,m}G_t^{k,m}}{c_{k,t}}
 f(y_{1:t}^{k,m})
\nonumber\\
&=\widehat Z_{k,t-1}\sum_{m=1}^M
\bar w_{t-1}^{k,m}G_t^{k,m}f(y_{1:t}^{k,m}).
\label{eq:island_unnormalized_update}
\end{align}
Thus the normalization factor $c_{k,t}$ cancels from the unnormalized island
measure. For a bounded $f$, define the one-step target update
\[
(\mathcal T_t f)(y_{<t})
:=\sum_{v\in\mathcal V_t(y_{<t})}
\frac{\varphi_t((y_{<t},v))}{\varphi_{t-1}(y_{<t})}
 f((y_{<t},v)).
\]
This function is bounded on the finite set of reachable states. Conditioning
on the complete population history and expanding the next-token draw yields
\begin{align}
&\mathbb E\!\left[
\widehat\Gamma_{k,t}^{M}(f)\mid\mathcal F_{t-1}
\right]
\nonumber\\
&=\widehat Z_{k,t-1}\sum_{m=1}^M\bar w_{t-1}^{k,m}
\sum_{v\in\mathcal V_t(y_{<t}^{k,m})}
q_t^{k,m}(v\mid\mathcal F_{t-1})G_t^{k,m}(v)
 f((y_{<t}^{k,m},v))
\nonumber\\
&=\widehat Z_{k,t-1}\sum_{m=1}^M\bar w_{t-1}^{k,m}
(\mathcal T_t f)(y_{<t}^{k,m}).
\label{eq:island_conditional_target_update}
\end{align}
The last equality uses \cref{eq:conditional_importance_identity}: the actual
proposal cancels with the proposal denominator in the incremental importance
ratio. Taking expectation again, applying the induction hypothesis to the
bounded function $\mathcal T_t f$, and then expanding its definition gives
\begin{align}
\mathbb E\!\left[\widehat\Gamma_{k,t}^{M}(f)\right]
&=\mathbb E\!\left[
\widehat Z_{k,t-1}\sum_{m=1}^M\bar w_{t-1}^{k,m}
(\mathcal T_t f)(y_{<t}^{k,m})
\right]
\nonumber\\
&=\Gamma_{t-1}(\mathcal T_t f)
\nonumber\\
&=\sum_{y_{<t}}\varphi_{t-1}(y_{<t})
\sum_{v\in\mathcal V_t(y_{<t})}
\frac{\varphi_t((y_{<t},v))}{\varphi_{t-1}(y_{<t})}
 f((y_{<t},v))
\nonumber\\
&=\sum_{y_{1:t}}\varphi_t(y_{1:t})f(y_{1:t})
=\Gamma_t(f).
\label{eq:island_unbiased_induction}
\end{align}
The first equality is the tower property together with
\cref{eq:island_conditional_target_update}; the second applies the induction
hypothesis at step $t-1$. In the final equality,
$\varphi_{t-1}(y_{<t})$ cancels and $(y_{<t},v)$ ranges over all step-$t$
states. The absorbing-EOS convention makes the same calculation valid for
completed responses.

If resampling is not triggered, the weighted empirical measure is unchanged.
If it is triggered, $\widehat Z_{k,t}$ remains unchanged and
\cref{eq:local_resampling_unbiased} gives
\[
\mathbb E\!\left[
\widehat Z_{k,t}\frac{1}{M}\sum_{m=1}^M
f(y_{1:t}^{k,A_{k,t}^m})
\middle|\mathcal F_t^-
\right]
=\widehat Z_{k,t}\sum_{m=1}^M
\bar w_t^{k,m}f(y_{1:t}^{k,m})
=\widehat\Gamma_{k,t}^{M}(f).
\]
Both cases therefore preserve the induction statement at step $t$, completing
the induction. The proof conditions on the joint population history, so the
adaptive proposal and resampling decisions are measurable and independence
among islands is not required.

Averaging \cref{eq:island_unbiasedness} over the $K$ islands gives
\cref{eq:pooled_unbiasedness}. Finally, take $f\equiv1$. Since the normalized
weights in each island sum to one,
\begin{equation}
\begin{aligned}
\mathbb E\!\left[\frac{1}{K}\sum_{k=1}^K\widehat Z_{k,t}\right]
&=\mathbb E\!\left[\widehat\Gamma_t^{K,M}(1)\right]\\
&=\Gamma_t(1)
=\sum_{y_{1:t}}\varphi_t(y_{1:t})=:Z_t.
\end{aligned}
\label{eq:normalizer_unbiasedness}
\end{equation}
This proves the normalizing-constant claim.

\end{proof}

\paragraph{Proposition 2 ($M\to\infty$ consistency).}
Fix $K$, a finite horizon $H$, and an input $(I,x)$. Before each next-token
draw, $\mathcal F_{t-1}$ determines the actual proposal $q_t^{k,m}$ used by
every particle. At routed steps, the proposal may depend on the particle
population observed at the preceding routing checkpoint.
Both the base and scout proposals are normalized over the
same finite vocabulary $\mathcal V$ and assign positive probability wherever
the corresponding target increment is positive. Exact score ties favor the
larger island index, then the larger within-island particle index, and finally
the larger region index. Under this deterministic ordering, the routing
decision and hence the actual proposal are determined by $\mathcal F_{t-1}$.
For the fixed input and finite $H$, the set of possible particle-local proposal
states is finite and independent of $M$. It comprises active and EOS-padded
response states, scout status, routed regions, and the cache states determined
by these finite histories. All proposal hyperparameters are also fixed
independently of $M$. Because each realized proposal is positive on the support
of its target increment, the minimum positive proposal probability over these
particle-local states is strictly positive.
The incremental importance weights therefore have uniformly bounded second moments: there exists a finite constant $C$, independent of
$M$ and the realized population history, such that
\begin{equation}
\mathbb E\!\left[\left(G_t^{k,m}(Y_t^{k,m})\right)^2\mid\mathcal F_{t-1}\right]\leq C,
\qquad t\leq H.
\label{eq:token_second_moment}
\end{equation}
Conditional on $\mathcal F_{t-1}$, the particles use independent random draws.
Let
\[
Z_t:=\sum_{y_{1:t}}\varphi_t(y_{1:t}),
\qquad
\pi_t:=\frac{\varphi_t}{Z_t}.
\]
At the terminal step, $\pi_H=\pi_\alpha^F$ and $Z_H=Z_\alpha^F$. Under these
conditions, for every bounded $f$,
\begin{equation}
\sum_{k=1}^K\sum_{m=1}^M
\widetilde W_H^{k,m}f(y^{k,m})
\xrightarrow[M\to\infty]{\mathsf P}\sum_{y}\pi_H(y)f(y).
\label{eq:pooled_consistency}
\end{equation}
Here $y^{k,m}$ denotes the completed terminal response of particle $(k,m)$.
In addition, the island normalizing-mass estimates satisfy
\[
\frac{1}{K}\sum_{k=1}^K\widehat Z_{k,H}
\xrightarrow[M\to\infty]{\mathsf P}
Z_H=Z_\alpha^F.
\]
\begin{proof}
We proceed by induction over token steps $t=0,\ldots,H$.

\noindent\textbf{Induction statements.}
For each fixed island $k$, after completing token step $t$ and any optional
resampling, we prove the following three statements.

\noindent\textbf{Statement 1 (empirical-measure consistency).} For every bounded test
function $f$ on the prefix space at step $t$,
\begin{equation}
\sum_{m=1}^M
\bar w_t^{k,m}f(y_{1:t}^{k,m})
\xrightarrow[M\to\infty]{\mathsf P}
\sum_{y_{1:t}}\pi_t(y_{1:t})f(y_{1:t}).
\label{eq:token_induction_hypothesis}
\end{equation}

\noindent\textbf{Statement 2 (weight control).} The squared normalized-weight mass
satisfies
\begin{equation}
S_{t,M}^k:=\sum_{m=1}^M(\bar w_t^{k,m})^2,
\qquad
S_{t,M}^k=O_{\mathsf P}(M^{-1}).
\label{eq:token_induction_weight_control}
\end{equation}

\noindent\textbf{Statement 3 (normalizing-mass consistency).} The island estimate
satisfies $\widehat Z_{k,t}\xrightarrow{\mathsf P}Z_t$.

\noindent\textbf{Base case ($t=0$).} Every particle represents the empty prefix and has
weight $1/M$, while $\pi_0$ is concentrated on that same empty prefix.
Therefore the two sides of Statement 1 are identical,
$S_{0,M}^k=M(1/M)^2=1/M$, and
$\widehat Z_{k,0}=Z_0=1$. All three statements hold at $t=0$.

\noindent\textbf{Inductive step ($t-1$ to $t$).} Assume all three statements hold at
step $t-1$. Let $\bar w_{t-1}^{k,m}$ be the normalized weight carried by
particle $(k,m)$ into token step $t$, after any optional resampling at step
$t-1$. We first establish the three statements after propagation and importance
weighting, before optional resampling at step $t$.

\noindent\textbf{(a) Shared corrected-propagation result.} At step $t$,
particle $(k,m)$ draws
$Y_t^{k,m}\sim q_t^{k,m}(\cdot\mid\mathcal F_{t-1})$. Write $G_t^{k,m}(v)$
for the incremental weight associated with candidate token $v$. After
sampling, the realized incremental weight is $G_t^{k,m}(Y_t^{k,m})$. For any
bounded test function $f$ on the prefix space at step $t$, define
$X_t^{k,m}:=G_t^{k,m}(Y_t^{k,m})
f((y_{<t}^{k,m},Y_t^{k,m}))$. Here $(y_{<t}^{k,m},v)$ denotes the length-$t$
state obtained by appending token $v$ to an active prefix. Under the
absorbing-EOS convention, it denotes the same completed response for a
completed state.
Conditioning on the complete population history gives
\begin{align}
&\mathbb E\!\left[X_t^{k,m}\mid\mathcal F_{t-1}\right]\nonumber\\
&=\sum_{v\in\mathcal V_t(y_{<t}^{k,m})}q_t^{k,m}(v\mid\mathcal F_{t-1})G_t^{k,m}(v)f((y_{<t}^{k,m},v))\nonumber\\
&=\sum_{v\in\mathcal V_t(y_{<t}^{k,m})}q_t^{k,m}(v\mid\mathcal F_{t-1})
\frac{\varphi_t((y_{<t}^{k,m},v))}
{\varphi_{t-1}(y_{<t}^{k,m})q_t^{k,m}(v\mid\mathcal F_{t-1})}
f((y_{<t}^{k,m},v))\nonumber\\
&=\sum_{v\in\mathcal V_t(y_{<t}^{k,m})}
\frac{\varphi_t((y_{<t}^{k,m},v))}{\varphi_{t-1}(y_{<t}^{k,m})}
f((y_{<t}^{k,m},v)).
\label{eq:adaptive_token_cancellation}
\end{align}
The population-dependent proposal therefore cancels from the conditional
target update. Consequently, the consistency argument does not require the
route-selected proposal itself to converge as $M$ grows.
Define the propagated weighted sum before normalization
\begin{equation}
U_{t,M}^k(f):=\sum_{m=1}^M\bar w_{t-1}^{k,m}G_t^{k,m}(Y_t^{k,m})
f((y_{<t}^{k,m},Y_t^{k,m})).
\label{eq:token_unnormalized_numerator}
\end{equation}

\noindent\textbf{Conditional mean.} By the definition of
$U_{t,M}^k(f)$, linearity of conditional expectation, and the proposal
cancellation above,
\begin{align}
&\mathbb E\!\left[U_{t,M}^k(f)\mid\mathcal F_{t-1}\right]
\nonumber\\
&=\mathbb E\!\left[
\sum_{m=1}^M\bar w_{t-1}^{k,m}G_t^{k,m}(Y_t^{k,m})
 f((y_{<t}^{k,m},Y_t^{k,m}))
\middle|\mathcal F_{t-1}
\right]
\nonumber\\
&=\sum_{m=1}^M\bar w_{t-1}^{k,m}
\mathbb E\!\left[
G_t^{k,m}(Y_t^{k,m})f((y_{<t}^{k,m},Y_t^{k,m}))
\mid\mathcal F_{t-1}
\right]
\nonumber\\
&=\sum_{m=1}^M\bar w_{t-1}^{k,m}
\sum_{v\in\mathcal V_t(y_{<t}^{k,m})}q_t^{k,m}(v\mid\mathcal F_{t-1})
G_t^{k,m}(v)f((y_{<t}^{k,m},v))
\nonumber\\
&=\sum_{m=1}^M\bar w_{t-1}^{k,m}
\sum_{v\in\mathcal V_t(y_{<t}^{k,m})}
\frac{\varphi_t((y_{<t}^{k,m},v))}
{\varphi_{t-1}(y_{<t}^{k,m})}
f((y_{<t}^{k,m},v)).
\label{eq:token_conditional_expectation}
\end{align}
The inner sum in the last line is a bounded function of the incoming prefix:
$f$ is bounded, $\mathcal V$ is finite, every reachable prefix has
$\varphi_{t-1}(y_{<t})>0$, and a fixed finite horizon admits only finitely many
reachable active and EOS-padded states. Statement 1 at step $t-1$ therefore
applies to this inner sum. Substituting
$\pi_{t-1}(y_{<t})=\varphi_{t-1}(y_{<t})/Z_{t-1}$ gives
\begin{align}
&\mathbb E\!\left[U_{t,M}^k(f)\mid\mathcal F_{t-1}\right]
\nonumber\\
&\xrightarrow{\mathsf P}
\sum_{y_{<t}}\pi_{t-1}(y_{<t})
\sum_{v\in\mathcal V_t(y_{<t})}
\frac{\varphi_t((y_{<t},v))}{\varphi_{t-1}(y_{<t})}
f((y_{<t},v))
\nonumber\\
&=\frac{1}{Z_{t-1}}
\sum_{y_{<t}}\sum_{v\in\mathcal V_t(y_{<t})}
\varphi_t((y_{<t},v))f((y_{<t},v))
\nonumber\\
&=\frac{\Gamma_t(f)}{Z_{t-1}}.
\label{eq:token_conditional_target_limit}
\end{align}
The second line substitutes the induction hypothesis. The following equality
uses the definition of $\pi_{t-1}$ and cancels
$\varphi_{t-1}(y_{<t})$. The double sum then equals $\Gamma_t(f)$ because
$(y_{<t},v)$ ranges over all length-$t$ prefixes.

\noindent\textbf{Vanishing propagation noise.} Conditional independence, the uniform
second-moment bound, and Statement 2 at step $t-1$ give
\begin{align}
&\operatorname{Var}\!\left[U_{t,M}^k(f)\mid\mathcal F_{t-1}\right]
\nonumber\\
&=\operatorname{Var}\!\left[
\sum_{m=1}^M\bar w_{t-1}^{k,m}X_t^{k,m}
\middle|\mathcal F_{t-1}
\right]
\nonumber\\
&=\sum_{m=1}^M(\bar w_{t-1}^{k,m})^2
\operatorname{Var}\!\left[X_t^{k,m}\mid\mathcal F_{t-1}\right]
\nonumber\\
&\leq\sum_{m=1}^M(\bar w_{t-1}^{k,m})^2
\mathbb E\!\left[(X_t^{k,m})^2\mid\mathcal F_{t-1}\right]
\nonumber\\
&=\sum_{m=1}^M(\bar w_{t-1}^{k,m})^2
\mathbb E\!\left[
\left(G_t^{k,m}(Y_t^{k,m})\right)^2
f((y_{<t}^{k,m},Y_t^{k,m}))^2
\mid\mathcal F_{t-1}
\right]
\nonumber\\
&\leq C\lVert f\rVert_\infty^2
\sum_{m=1}^M(\bar w_{t-1}^{k,m})^2
=C\lVert f\rVert_\infty^2S_{t-1,M}^k
=O_{\mathsf P}(M^{-1}).
\label{eq:adaptive_token_variance}
\end{align}
Here $\lVert f\rVert_\infty$ is the largest absolute value of $f$ over
length-$t$ prefixes and is finite because $f$ is bounded. For any
$\varepsilon>0$, conditional Chebyshev's inequality gives
\begin{align}
&\Pr\!\left(
\left|U_{t,M}^k(f)-\mathbb E[U_{t,M}^k(f)\mid\mathcal F_{t-1}]\right|
>\varepsilon
\middle|\mathcal F_{t-1}
\right)
\nonumber\\
&\quad\leq
\frac{\operatorname{Var}[U_{t,M}^k(f)\mid\mathcal F_{t-1}]}
{\varepsilon^2}
\leq
\frac{C\lVert f\rVert_\infty^2}{\varepsilon^2}S_{t-1,M}^k
\xrightarrow{\mathsf P}0.
\label{eq:token_conditional_chebyshev}
\end{align}
These conditional probabilities lie in $[0,1]$. Taking expectations and using
the law of total probability therefore gives
\begin{equation}
U_{t,M}^k(f)-\mathbb E\!\left[U_{t,M}^k(f)\mid\mathcal F_{t-1}\right]
\xrightarrow{\mathsf P}0.
\label{eq:token_centered_convergence}
\end{equation}
Thus the conditional mean approaches the correct target update, and the random
fluctuation around that mean vanishes.

Combining \cref{eq:token_centered_convergence} and
\cref{eq:token_conditional_target_limit} makes both contributions explicit:
\begin{align}
U_{t,M}^k(f)
&=\left(
U_{t,M}^k(f)-\mathbb E[U_{t,M}^k(f)\mid\mathcal F_{t-1}]
\right)
+\mathbb E[U_{t,M}^k(f)\mid\mathcal F_{t-1}]
\nonumber\\
&\xrightarrow{\mathsf P}
0+\frac{\Gamma_t(f)}{Z_{t-1}}
=\frac{\Gamma_t(f)}{Z_{t-1}}.
\label{eq:token_target_update}
\end{align}

\noindent\textbf{(b) Proof of Statement 1: empirical-measure consistency.}
Taking \(f=1\) in the shared propagation limit above gives
\begin{equation}
c_{k,t}:=U_{t,M}^k(1)
\xrightarrow{\mathsf P}\frac{Z_t}{Z_{t-1}}>0.
\label{eq:token_normalizer_increment_limit}
\end{equation}
The propagated normalized weights satisfy
\(\bar w_t^{k,m}=\bar w_{t-1}^{k,m}G_t^{k,m}(Y_t^{k,m})/c_{k,t}\).
Therefore,
\begin{align}
\sum_{m=1}^M\bar w_t^{k,m}f(y_{1:t}^{k,m})
&=\frac{U_{t,M}^k(f)}{c_{k,t}}
\nonumber\\
&\xrightarrow{\mathsf P}
\frac{\Gamma_t(f)/Z_{t-1}}{Z_t/Z_{t-1}}
=\sum_{y_{1:t}}\pi_t(y_{1:t})f(y_{1:t}).
\label{eq:token_empirical_measure_propagation}
\end{align}
The numerator uses \cref{eq:token_target_update}, and the positive denominator
uses \cref{eq:token_normalizer_increment_limit}. This proves Statement 1 at
step \(t\) before optional resampling.

\noindent\textbf{(c) Proof of Statement 2: weight control.}
The uniform second-moment bound gives
\begin{align}
&\mathbb E\!\left[
\sum_{m=1}^M(\bar w_{t-1}^{k,m})^2
\left(G_t^{k,m}(Y_t^{k,m})\right)^2
\middle|\mathcal F_{t-1}
\right]
\nonumber\\
&\quad\leq
C\sum_{m=1}^M(\bar w_{t-1}^{k,m})^2
=O_{\mathsf P}(M^{-1}).
\label{eq:token_weight_second_moment}
\end{align}
Conditional Markov's inequality shows that the sum inside the expectation is
itself \(O_{\mathsf P}(M^{-1})\). Dividing it by \(c_{k,t}^2\), whose
limit is positive, gives
\begin{equation}
\sum_{m=1}^M(\bar w_t^{k,m})^2=O_{\mathsf P}(M^{-1}).
\label{eq:token_weight_propagation}
\end{equation}
This proves Statement 2 at step \(t\) before optional resampling.

\noindent\textbf{(d) Proof of Statement 3: normalizing-mass consistency.}
By Statement 3 at step \(t-1\), the island-mass recursion, and
\cref{eq:token_normalizer_increment_limit},
\begin{equation}
\widehat Z_{k,t}
=\widehat Z_{k,t-1}c_{k,t}
\xrightarrow{\mathsf P}
Z_{t-1}\frac{Z_t}{Z_{t-1}}=Z_t.
\label{eq:token_normalizer_consistency_step}
\end{equation}
This proves Statement 3 at step \(t\) before optional resampling.

\noindent\textbf{(e) Preservation of Statements 1--3 under optional resampling.}
If resampling is not triggered, the particles and normalized weights are
carried forward unchanged, so all three statements remain true. If it is
triggered, stratified resampling draws island-local ancestors
\(A_{k,t}^{1:M}\) and assigns weight \(1/M\) to every descendant.
Conditional on the pre-resampling population \(\mathcal F_t^-\),
\begin{align}
\mathbb E\!\left[
\frac{1}{M}\sum_{m=1}^M f(y_{1:t}^{k,A_{k,t}^m})
\middle|\mathcal F_t^-
\right]
&=\sum_{j=1}^M\bar w_t^{k,j}f(y_{1:t}^{k,j}),
\nonumber\\
\operatorname{Var}\!\left[
\frac{1}{M}\sum_{m=1}^M f(y_{1:t}^{k,A_{k,t}^m})
\middle|\mathcal F_t^-
\right]
&\leq\frac{\lVert f\rVert_\infty^2}{M}.
\label{eq:stratified_induction_mean_variance}
\end{align}
\textbf{Statement 1} is preserved because the first line is conditionally
unbiased and the second, together with Chebyshev's inequality, makes the
resampling error converge to zero in probability. \textbf{Statement 2} is
preserved because the equal descendant weights have squared mass exactly
\(1/M\). \textbf{Statement 3} is preserved because resampling leaves
\(\widehat Z_{k,t}\) unchanged. The ESS trigger is determined by
\(\mathcal F_t^-\), so the argument covers both branches without requiring
the trigger to stabilize as \(M\) grows. This completes the inductive step from \(t-1\) to \(t\), and hence the
induction over \(t=0,\ldots,H\).

Set \(t=H\). Statements 1 and 2 give, for each fixed island \(k\),
\begin{equation}
\sum_{m=1}^M\bar w_H^{k,m}f(y^{k,m})
\xrightarrow{\mathsf P}\sum_y\pi_H(y)f(y),
\qquad
\sum_{m=1}^M(\bar w_H^{k,m})^2=O_{\mathsf P}(M^{-1}).
\label{eq:per_island_terminal_consistency}
\end{equation}
Statement 3 gives
\(\widehat Z_{k,H}\xrightarrow{\mathsf P}Z_H\).

Finally, substitute
\(\widetilde W_H^{k,m}=\widehat Z_{k,H}\bar w_H^{k,m}/
\sum_{j=1}^K\widehat Z_{j,H}\). Because \(K\) is fixed, the per-island limits
above imply
\begin{align}
\sum_{k=1}^K\sum_{m=1}^M\widetilde W_H^{k,m}f(y^{k,m})
&=\frac{\sum_{k=1}^K\widehat Z_{k,H}
\sum_{m=1}^M\bar w_H^{k,m}f(y^{k,m})}
{\sum_{j=1}^K\widehat Z_{j,H}}
\nonumber\\
&\xrightarrow{\mathsf P}
\frac{\sum_{k=1}^K Z_H\sum_y\pi_H(y)f(y)}{KZ_H}
=\sum_y\pi_H(y)f(y).
\label{eq:pooled_consistency_derivation}
\end{align}
This proves \cref{eq:pooled_consistency}. The same normalizer limit gives
\(K^{-1}\sum_{k=1}^K\widehat Z_{k,H}\xrightarrow{\mathsf P}Z_H=Z_\alpha^F\),
which completes the proof.

\end{proof}

\section{Answer-marginal power readout}
\label{app:answer_power}
For the population-level analysis, write
$\pi_\alpha(y)=\pi_\alpha^F(y\mid I,x)$ and suppress the fixed conditioning
pair $(I,x)$. Let $\operatorname{Ans}:\mathcal Y_H\rightarrow\mathcal A$ map a completed
response to its canonical answer. The sequence-power target induces
\begin{equation}
\mu_\alpha(a)
=\sum_{y:\operatorname{Ans}(y)=a}\pi_\alpha(y),
\label{eq:exact_answer_marginal}
\end{equation}
and the corresponding finite answer power is
\begin{equation}
q_{\alpha,\gamma}(a)
=\frac{\mu_\alpha(a)^\gamma}
{\sum_b\mu_\alpha(b)^\gamma},
\qquad \gamma\geq1.
\label{eq:exact_answer_power}
\end{equation}
These population quantities are approximated by
\cref{eq:empirical_answer_marginal,eq:empirical_answer_power}.

After drawing $a^\star\sim\widehat q_{\alpha,\gamma}$, the
implementation returns a complete response by drawing one supporting terminal
particle according to
\begin{equation}
\Pr(k,m\mid a^\star)
=\frac{\widetilde W_H^{k,m}\mathbf 1[\operatorname{Ans}(y^{k,m})=a^\star]}
{\widehat\mu_\alpha(a^\star)}.
\label{eq:conditional_supporting_trajectory}
\end{equation}
For a terminal particle $(k,m)$, let $a_{k,m}:=\operatorname{Ans}(y^{k,m})$. This particle can be returned only when the first-stage answer draw selects $a_{k,m}$. Therefore,\begin{align}\Pr(k,m)&=\Pr(a^\star=a_{k,m})  \Pr(k,m\mid a^\star=a_{k,m})\nonumber\\&=\frac{\widehat\mu_\alpha(a_{k,m})^\gamma}{\sum_b\widehat\mu_\alpha(b)^\gamma}\frac{\widetilde W_H^{k,m}}{\widehat\mu_\alpha(a_{k,m})}.\label{eq:two_stage_particle_derivation}\end{align}The indicator in \cref{eq:conditional_supporting_trajectory} equals one by the definition of $a_{k,m}$. Cancelling one factor of $\widehat\mu_\alpha(a_{k,m})$ gives the
induced terminal-particle distribution
\begin{equation}
\widehat\pi_{\alpha,\gamma}(k,m)
=\frac{\widetilde W_H^{k,m}
\widehat\mu_\alpha(\operatorname{Ans}(y^{k,m}))^{\gamma-1}}
{\sum_b\widehat\mu_\alpha(b)^\gamma}.
\label{eq:two_stage_particle_distribution}
\end{equation}

The population analogue replaces the weighted terminal-particle measure by $\pi_\alpha$. For a completed trajectory $y$, let $a_y:=\operatorname{Ans}(y)$. At the population level, this gives
\begin{equation}
\begin{aligned}\pi_{\alpha,\gamma}^{\mathrm{2stage}}(y)&:=q_{\alpha,\gamma}(a_y)\pi_\alpha(y\mid\operatorname{Ans}(Y)=a_y)\\&=\frac{\mu_\alpha(a_y)^\gamma}{\sum_b\mu_\alpha(b)^\gamma}\frac{\pi_\alpha(y)}{\mu_\alpha(a_y)}\\&
=\frac{\pi_\alpha(y)
\mu_\alpha(\operatorname{Ans}(y))^{\gamma-1}}
{\sum_b\mu_\alpha(b)^\gamma}.
\end{aligned}\label{eq:population_two_stage_target}
\end{equation}

Empirically, \cref{eq:empirical_answer_marginal} defines the
pushforward of the weighted terminal particle measure through $\operatorname{Ans}$. Both distributions recover their respective first-stage terminal measures when $\gamma=1$. For $\gamma>1$, trajectories supporting the same answer receive the same answer-mass multiplier, preserving their relative probabilities within that answer while favoring answers with larger marginal mass. This empirical readout does not alter the unbiased normalizing-constant statement in
\cref{eq:pooled_unbiasedness}.

\subsection{Split sharpening and cross-trajectory support}
\label{app:split_sharpening}
For an answer $a$, enumerate its trajectories as
$\mathcal Y_a=\{y_{a,1},y_{a,2},\ldots\}$ and write
$p_{a,i}=p_F(y_{a,i}\mid I,x)$. Common normalizing constants can be omitted
when comparing answers. Applying one sequence exponent $\alpha\gamma$ gives
the answer mass
\begin{equation}
D_{\alpha\gamma}(a)=\sum_i p_{a,i}^{\alpha\gamma},
\label{eq:direct_high_power_mass}
\end{equation}
whereas splitting the powers across trajectories and answers gives
\begin{equation}
S_{\alpha,\gamma}(a)
=\left(\sum_i p_{a,i}^{\alpha}\right)^\gamma.
\label{eq:split_power_mass}
\end{equation}
For $\gamma=2$,
\begin{equation}
S_{\alpha,2}(a)
=D_{2\alpha}(a)+2\sum_{i<j}p_{a,i}^{\alpha}p_{a,j}^{\alpha}.
\label{eq:split_power_general_cross_terms}
\end{equation}
The excess over direct sequence power is therefore exactly the sum of
same-answer cross terms. It is nonnegative and is zero only when at most one
trajectory with positive mass supports the answer. Because this excess differs
across answers, normalization can change their ranking. In particular,
$(\alpha,\gamma)=(2,2)$ retains the direct fourth-power term while also
crediting collective support from multiple trajectories.

\paragraph{A separating two-token example.}
Following the two-token construction used to illustrate sequence-power
lookahead~\citep{karan2025reasoning}, let the first token choose a reasoning
branch and the second choose a completion within that branch. Consider six
complete trajectories with
\begin{equation}
\begin{aligned}
p(a_1)=p(a_2)=p(a_3)&=3/22,\\
p(b_1)=p(b_2)&=4/22,\\
p(c_1)&=5/22,
\end{aligned}
\label{eq:two_token_separating_example}
\end{equation}
where branches $a$, $b$, and $c$ yield answers $A$, $B$, and $C$,
respectively. This is a valid autoregressive model: the first-token
probabilities are $(9,8,5)/22$, and completions are uniform within each branch.
Under sequence-level $\alpha=4$, the first-token and answer masses are obtained
by summing the fourth powers of complete future trajectories:
\begin{equation}
D_4(A):D_4(B):D_4(C)
=3\cdot3^4:2\cdot4^4:5^4
=243:512:625.
\label{eq:two_token_direct_fourth_power}
\end{equation}
Thus weak continuations are suppressed before the first token is sampled.
With the split construction, the sequence-square marginal is
$\mu_2(A):\mu_2(B):\mu_2(C)=27:32:25$, and the answer-level square gives
$q_{2,2}(A):q_{2,2}(B):q_{2,2}(C)=729:1024:625$.

\begin{table}[!htbp]
\caption{A two-token autoregressive example in which base sampling, split
sharpening, and direct fourth-power sequence sampling have different answer
modes.}
\label{tab:split_power_example}
\begin{center}
\scriptsize
\setlength{\tabcolsep}{3pt}
\begin{tabular}{@{}lccc@{}}
\toprule
& Base $\alpha=1$ & Split $(2,2)$ & Seq. $\alpha=4$ \\
\midrule
$A$ (3 weak) & $9/22$ & $729/2378$ & $243/1380$ \\
$B$ (2 moderate) & $8/22$ & $1024/2378$ & $512/1380$ \\
$C$ (1 strong) & $5/22$ & $625/2378$ & $625/1380$ \\
\midrule
Mode & $A$ & $B$ & $C$ \\
\bottomrule
\end{tabular}
\end{center}
\end{table}

Base sampling favors $A$ because its three individually weakest trajectories
carry the largest total mass. Direct sequence-level fourth power selects $C$,
whose powered future mass is already largest at the first token. Split
sharpening selects $B$: the trajectory square discounts weak paths while
retaining collective support from two moderate paths, and the second square is
applied only after their answer mass is combined.

The split therefore remains fourth-order in the base trajectory probabilities
while avoiding both unweighted majority behavior and the pathwise concentration
of direct fourth-power sequence sampling.

\section{Finite-Particle Behavior on Direct-Answer Tasks}
\label{app:direct_answer_finite_particles}

MMStar-P consists of single-choice questions, and our instruction asks the model to end with \texttt{Final answer:...}. Across four seeds, 1,854 of 2,000 selected outputs under $(\alpha,\gamma)=(2,2)$ and 1,866 of 2,000 under $(4,1)$ contain this marker. All 3,060 selected RealWorldQA outputs give a direct answer. These formats motivate a simplified analysis in which one generated token determines the answer.

\paragraph{Direct-answer model.}
We assume that each admissible answer $a$ corresponds to one complete trajectory. The answer is either the first generated token or the token immediately following the fixed generated prefix \texttt{Final answer:}. We assume one fixed format for each question within a dataset, so the two cases never occur together in one particle population. Under the Qwen tokenization used here, \texttt{Final answer:} comprises the tokens \texttt{Final}, \texttt{ answer}, and \texttt{:}; an option such as \texttt{ A} is the next token. Thus the answer position is $t_\ast=4$ in the fixed-prefix format and $t_\ast=1$ in the immediate-answer format. We assume that the remaining suffix and termination are fixed and contribute the same likelihood factor for every answer. Particle draws are independent, without scouting or resampling.

Let $p_a$ be the base-model probability of selecting answer $a$, with $\sum_a p_a=1$. At the answer position, the run with sequence exponent $\alpha$ draws from
\begin{equation}
Q_\alpha(a)=\frac{p_a^{\beta_{t_\ast}^{(\alpha)}}}{\sum_b p_b^{\beta_{t_\ast}^{(\alpha)}}},
\qquad
\beta_t^{(\alpha)}=1+(\alpha-1)\min\{t/128,1\}.
\label{eq:direct_answer_proposal}
\end{equation}
For $t_\ast=1$, the proposal exponents are $\beta_1^{(2)}=1.0078125$ and $\beta_1^{(4)}=1.0234375$. For $t_\ast=4$, they are $\beta_4^{(2)}=1.03125$ and $\beta_4^{(4)}=1.09375$. These exponents determine the proposals, not the terminal targets. Importance correction and the updates after completion bring each trajectory to terminal exponent $\alpha$. Since the other trajectory factors are common to all answers and cancel on normalization, its terminal importance weight is proportional to
\begin{equation}
W_\alpha(a)=\frac{p_a^\alpha}{Q_\alpha(a)}.
\label{eq:direct_answer_weight}
\end{equation}

\paragraph{Finite-particle readouts.}
Let $n=KM=32$ and let $N_a^{(\alpha)}$ be the number of particles selecting $a$ in a run with exponent $\alpha$. Direct sequence power $(4,1)$ assigns $a$ the probability
\begin{equation}
\widehat P_{4,1}(a)=
\frac{N_a^{(4)}W_4(a)}{\sum_b N_b^{(4)}W_4(b)}.
\label{eq:direct_answer_readout_four}
\end{equation}
The two-stage $(2,2)$ readout first aggregates particles by answer and then squares each aggregate:
\begin{equation}
\widehat P_{2,2}(a)=
\frac{\bigl(N_a^{(2)}W_2(a)\bigr)^2}{\sum_b\bigl(N_b^{(2)}W_2(b)\bigr)^2}.
\label{eq:direct_answer_readout_split}
\end{equation}
As $n\to\infty$, $N_a^{(\alpha)}/n\to Q_\alpha(a)$. Both readouts therefore converge to $p_a^4/\sum_b p_b^4$. At finite $n$, however, the second-stage power acts on an estimated answer mass. Since $N_a^{(2)}\sim\operatorname{Binomial}(n,Q_2(a))$, its mean is $nQ_2(a)$ and its variance is $nQ_2(a)(1-Q_2(a))$. Applying $\mathbb E[N^2]=\operatorname{Var}(N)+\mathbb E[N]^2$ gives
\begin{align}
\mathbb E\!\left[\bigl(N_a^{(2)}W_2(a)\bigr)^2\right]
&=\frac{p_a^4}{Q_2(a)^2}\left[nQ_2(a)(1-Q_2(a))+n^2Q_2(a)^2\right]\nonumber\\
&=n^2p_a^4+np_a^4\left(\frac{1}{Q_2(a)}-1\right).
\label{eq:direct_answer_mass_second_moment}
\end{align}
The second term comes from finite-particle count variation. To obtain expected pass@1, the answer masses must be normalized within each realized population before averaging over populations.

\paragraph{Exact finite-particle example.}
Consider four answers with $p=(0.4,0.3,0.2,0.1)$, where the highest-probability answer $A$ is correct. A population is described by nonnegative counts $\mathbf n=(n_A,n_B,n_C,n_D)$ summing to 32. There are $\binom{32+3}{3}=\binom{35}{3}=6{,}545$ such vectors: distributing 32 selections among four answers is equivalent to placing three separators among 35 positions. Under proposal $Q_\alpha$, the probability of a count vector is
\begin{equation}
\Pr_\alpha(\mathbf n)=
\frac{32!}{n_A!n_B!n_C!n_D!}
\prod_{a\in\{A,B,C,D\}}Q_\alpha(a)^{n_a}.
\label{eq:direct_answer_count_probability}
\end{equation}
For each vector, we insert its counts into \cref{eq:direct_answer_readout_four,eq:direct_answer_readout_split} to obtain the conditional probability of selecting $A$, multiply by \cref{eq:direct_answer_count_probability}, and sum over all 6,545 vectors:
\begin{equation}
\mathbb E[\widehat P_{\alpha,\gamma}(A)]
=\sum_{\substack{\mathbf n\geq0\\n_A+n_B+n_C+n_D=32}}
\Pr_\alpha(\mathbf n)\,\widehat P_{\alpha,\gamma}(A\mid\mathbf n),
\quad(\alpha,\gamma)\in\{(4,1),(2,2)\}.
\label{eq:direct_answer_expected_pass1}
\end{equation}
The two runs use their respective proposals $Q_4$ and $Q_2$. Evaluating this finite sum gives \cref{tab:direct_answer_finite_particles}.

\begin{table}[!htbp]
\caption{Expected pass@1 in the 32-particle direct-answer example with $p=(0.4,0.3,0.2,0.1)$ and correct answer $A$.}
\label{tab:direct_answer_finite_particles}
\begin{center}
\small
\begin{tabular}{lcc}
\toprule
Answer position & Direct $(4,1)$ & Two-stage $(2,2)$ \\
\midrule
First generated token, $t_\ast=1$ & $71.59\%$ & $69.12\%$ \\
After \texttt{Final answer:}, $t_\ast=4$ & $71.62\%$ & $69.14\%$ \\
\bottomrule
\end{tabular}
\end{center}
\end{table}

As $n$ grows, both readouts converge to $0.4^4/(0.4^4+0.3^4+0.2^4+0.1^4)=72.32\%$. This idealized direct-answer case illustrates a finite-particle mechanism relevant to MMStar-P and RealWorldQA: when each answer has only one trajectory, aggregation supplies no cross-trajectory support, while the $(2,2)$ readout squares noisy empirical answer masses. The exact calculation shows how direct $(4,1)$ can therefore have higher expected pass@1 in this setting.

\section{Inference algorithm}
\label{app:algorithm}
\Cref{alg:resight_smc} summarizes the complete inference procedure, from island-local SMC and visual scouting to answer-marginal sampling.

\begin{algorithm*}[t]
\caption{ReSight-SMC}
\label{alg:resight_smc}
\begin{algorithmic}[1]
\Require Image $I$, prompt $x$, horizon $H$, islands $K$, particles $M$
\State Initialize $KM$ base-proposal particles and a multiscale region bank
\State Cache the image-token keys at the routing layer
\State Set the active scout set $\mathcal S\gets\varnothing$
\For{$t=1,\ldots,H$}
  \ForAll{unfinished particles $(k,m)$}
    \If{$(k,m)\in\mathcal S$}
      \State Sample the next token from $q_t^{A,k,m}$ in
        \cref{eq:attention_proposal}
      \State Update the base-model likelihood and importance weight
      \If{the sampled token is nonterminal}
        \State Advance the persistent base target state with that token
        \If{another scout proposal is required}
          \State Advance the temporary attention state with that token
        \EndIf
      \EndIf
    \Else
      \State Sample the next token from $q_t^F$
      \State Update the base-model likelihood and importance weight
      \If{the sampled token is nonterminal}
        \State Advance the persistent base target state with that token
      \EndIf
    \EndIf
  \EndFor
  \State Advance the bridge for completed trajectories
  \State Normalize weights and update $\widehat Z_{k,t}$ within each island
  \If{$t=\tau_{\mathrm{vis}}+L_{\mathrm{vis}}$}
    \State Release the temporary attention states and set $\mathcal S\gets\varnothing$
  \EndIf
  \If{$t$ is an SMC checkpoint}
    \For{$k=1,\ldots,K$}
      \State Compute $\ESS_{k,t}$
      \If{$\ESS_{k,t}<\rho_wM$}
        \State Resample only island $k$ and reset its weights to $1/M$
      \EndIf
    \EndFor
  \EndIf
  \If{$t=\tau_{\mathrm{vis}}$}
    \State Compute the per-island quotas $B_k$ and total budget $B$ from
      \cref{eq:scout_budget}
    \State Score eligible particle--region pairs using
      \cref{eq:particle_region_utility}
    \State Select $B$ pairs using \cref{eq:greedy_route}
    \State Fork their decoder states, replay $y_t$ under
      \cref{eq:attention_bias}, and set $\mathcal S$ to the selected particles
  \EndIf
\EndFor
\State Form terminal masses $\widetilde W_H^{k,m}$ using
  \cref{eq:island_terminal_weight}
\State Aggregate $\widehat\mu_\alpha$ by canonical answer using
  \cref{eq:empirical_answer_marginal}
\State Draw $a^\star\sim\widehat q_{\alpha,\gamma}$ using
  \cref{eq:empirical_answer_power}
\State Draw a supporting particle conditional on $a^\star$ using
  \cref{eq:conditional_supporting_trajectory} and return its response
\end{algorithmic}
\end{algorithm*}

\section{Computational cost}
\label{app:complexity}
\Cref{tab:system_cost} reports measured end-to-end latency and peak memory for base sampling, Power-SMC, and \method under a common hardware and software setup.

\begin{table}[!htbp]
\caption{End-to-end inference cost with Qwen2.5-VL-7B-Instruct on one RTX
5090. Latency and peak memory are question-weighted over LogicVista, MathVista,
and MMStar-R for seed 0. All systems use the Transformers backend.}
\label{tab:system_cost}
\begin{center}
\scriptsize
\setlength{\tabcolsep}{5pt}
\begin{tabular}{lccc}
\toprule
System & Population & Latency (s) & Peak VRAM (GB)\\
\midrule
Base sampling & 1 & 4.11 & 15.61\\
Power-SMC & $1\times32$ & 7.85 & 17.27\\
\method & $4\times8$ & 10.37 & 17.35\\
\bottomrule
\end{tabular}
\end{center}
\end{table}

In the worst case, when all $KM$ trajectories remain active for $H$ decoding steps, Island-SMC requires
\begin{equation}
T_F=\mathcal O(KMH C_F^{\mathrm{step}}).
\end{equation}
The visual phase creates at most $B$ temporary scout decoder
states. Let $N_I:=|\mathcal V_I|$ be the number of original-image visual tokens, $N_h$ the number
of heads, $d_h$ the per-head key dimension, and $C_F^{\mathrm{step}}$ one
full-context decoder step. Its additional computation is
\begin{equation}
T_A=\mathcal O\!\left(
KMN_hN_Id_h
+BKMG
+B(L_{\mathrm{vis}}+1)C_F^{\mathrm{step}}
\right).
\label{eq:visual_complexity}
\end{equation}
The first term computes prefix-to-image relevance at the routing layer, the
second performs greedy routing, and the final term covers one-token replay plus
the bounded scout episode. Image keys and particle caches come from the initial
prefill, and all temporary states are released after the visual phase.

Additional peak state memory is
\begin{equation}
S_A=\mathcal O\!\left(
B L(P_F+\tau_{\mathrm{vis}}+L_{\mathrm{vis}})d_{\mathrm{kv}}+BN_I
\right),
\end{equation}
where $L$ is the number of decoder layers, $P_F$ the original multimodal-prefix
length, and $d_{\mathrm{kv}}$ the per-layer key--value width. If $U\leq KM$
distinct canonical answers occur, terminal
aggregation and sampling require $\mathcal O(KM)$ time and $\mathcal O(U)$
memory, with no additional model forward.
\section{Experimental details}
\label{app:additional}
The main configuration uses $H=1024$, $\alpha=2$, and
\begin{equation}
\beta_t=1+(\alpha-1)\min\{t/128,1\},
\qquad \beta_H=\alpha.
\end{equation}
The base sampler uses temperature $1$. The low-temperature baseline draws one
autoregressive response with temperature $1/\alpha=0.5$ and otherwise uses the same prompt,
image preprocessing, response horizon, and full-vocabulary sampling rule. Both
use $\operatorname{top\_p}=1$ and $\operatorname{top\_k}=0$. The low-temperature
baseline does not apply sequence-level importance weights or resampling.
Power-SMC uses one population of 32 particles with global resampling.
The island-only variant and \method use $K=4$ islands with $M=8$ particles, stratified
resampling with threshold $\rho_w=0.5$, and checkpoint interval
$L_{\mathrm{SMC}}=32$. Every triggered resampling operation is confined to one
island.

The visual configuration is
\begin{equation}
\rho_V=0.25,\quad
\tau_{\mathrm{vis}}=40,\quad
L_{\mathrm{vis}}=16,\quad
\lambda_I=\log 2,\quad
\lambda_R=\log 4.
\end{equation}
Thus the default population assigns two scouts to each fully active island, for
at most eight scouts in total, and retains base-proposal anchors in every island.
Visual scouting is invoked once after token 40: tokens 41--56 use the scout proposals,
and tokens 57--64 return to
the base proposal before the token-64 resampling checkpoint. The region
bank contains $2\times2$ and $3\times3$ grid cells together with horizontal and
vertical thirds. We set the area exponent to $\zeta=0.75$ and the overlap
coefficient to $\mu=1$. Routing uses the final decoder layer. The attention
bias is applied at every decoder layer and shared across heads.
With these values, attention odds for every image token are doubled relative to
unmodified keys, and tokens inside the routed region receive a further factor
of four.

We fix prompting by benchmark category. MathVista, MMStar-R, and LogicVista use
CoT prompting. The perception-focused MMStar-P and RealWorldQA use direct
answering. All five benchmarks use $\tau_{\mathrm{vis}}=40$ and retain the same
scout fraction, episode length, attention biases, region bank, and
particle budget. A response that terminates before token 40 simply skips the
scout stage. 

The main \method configuration uses answer exponent $\gamma=2$. Power-SMC,
the island-only variant, and the trajectory-stage ReSight ablation use $\gamma=1$.
Terminal answers are grouped by the reference-free canonicalizer detailed in
\cref{app:prompt_eval}. The returned reasoning text comes from a terminal particle that
supports the sampled canonical answer.

The component ablation compares $K=1,M=32$ with $K=4,M=8$ under identical
base proposals, bridge schedules, ESS thresholds, and total particle
counts. It then adds the complete visual proposal stage and the answer-marginal
readout in turn. 

The backbone suite contains Qwen2.5-VL-3B-Instruct,
Qwen2.5-VL-7B-Instruct, Qwen3-VL-4B-Instruct, and
Qwen3-VL-8B-Instruct. The Qwen2.5-VL post-training references are
\texttt{maveryn/trace-qwen2.5-vl-3b} and
\texttt{maveryn/trace-qwen2.5-vl-7b}~\citep{alam2026trace}, trained with GRPO on
the 64K-instance Trace corpus. At 7B we additionally evaluate
Game-RL-Qwen2.5-VL-7B~\citep{tong2025code2logic}, trained with GRPO on GameQA. Each checkpoint uses the same
benchmark-specific prompt, response horizon, image preprocessing, and answer
evaluator as its corresponding training-free backbone. Ablations and appendix
analyses use Qwen2.5-VL-7B-Instruct.

\subsection{Prompts and answer evaluation}
\label{app:prompt_eval}

\paragraph{Prompt protocol.}
All compared methods use each checkpoint's native chat template without an additional system message.
The
user turn contains the image, the benchmark-released question, and, where
applicable, a fixed generation instruction.  MathVista, LogicVista, and
MMStar-R append
\verb+Think step by step and end with `Final answer: ...`.+, whereas
MMStar-P appends
\verb+Answer directly and end with `Final answer: ...`.+.  RealWorldQA
already includes a direct-answer instruction in its released question, so no
additional suffix is appended.  Base sampling, low-temperature sampling,
Power-SMC, \method, and the post-trained checkpoints use the same
dataset-specific prompt.

\paragraph{Deterministic answer extraction.}
The shared parser applies the following priority order: the last output line
that parses as a JSON object with an \texttt{answer} field; the last span
introduced by \texttt{Final answer:}, \texttt{Answer:}, or
their Chinese equivalent; the last \verb|\boxed{...}| expression; and finally the last
nonempty output line.  Textual comparisons apply Unicode NFKC normalization,
lowercasing, Unicode-minus normalization, removal of common surrounding
punctuation and LaTeX \texttt{\textbackslash text\{\}} or
\texttt{\textbackslash mathrm\{\}} wrappers, and 
whitespace normalization.
No LLM judge or LLM-based answer extractor is used.

\paragraph{Dataset-specific scoring.}
We use the 1,500 questions in the \texttt{Lin-Chen/MMStar} \texttt{val} split and partition them by the released \texttt{category} field. MMStar-R contains instance reasoning, logical reasoning, math, and science and technology; MMStar-P contains coarse perception and fine-grained perception (250 questions per category). The subsets are disjoint and cover the full split. Cross-subset comparisons reflect both category and prompting differences.

For MathVista, we use the 1,000-example \texttt{testmini} split and
answer-type-aware normalization.  Multiple-choice predictions are mapped from
an option label or generated option text to the listed choice; integer answers
are converted numerically, floating-point answers are rounded to the
dataset-provided precision, and list and other free-form answers use normalized
exact match.  This follows MathVista's answer-type-aware normalization while
replacing its optional LLM answer-extraction stage with the deterministic
parser above.  MMStar-R and MMStar-P use the same option-letter exact-match
evaluator,
and a generated option string is mapped back to its listed letter.

RealWorldQA~\citep{xai2024realworldqa} uses its released 765-question test set,
which combines multiple-choice and short-answer questions over real-world
scenes. We retain the released question text and direct-answer
instructions. Multiple-choice responses are scored by extracting the option
letter; other responses use case-insensitive exact match after removing a final
period. The visual ablation uses this protocol for both rows.

LogicVista~\citep{xiao2024logicvista} uses its released 448-question test split
and reports accuracy over the five reasoning skills. We extract the final option
labels and require exact equality with the reference set; answer order is
ignored for multi-select questions.

\paragraph{Reference-free answer grouping.}
Before answer-marginal aggregation, the same deterministic extraction rules
canonicalize every terminal particle.  Canonicalization may use the dataset
identifier, question type, and listed choices, but it never accesses the
reference answer.  Terminal masses are grouped by these canonical strings,
powered, and sampled as described in \cref{app:answer_power}. Only the sampled
answer is then passed to the benchmark-specific evaluator.

\paragraph{Numerical evaluation.}
Proposal distributions and target factors are formed by casting logits to
FP32 before a full-vocabulary log-softmax. Incremental and cumulative weights remain in the log domain.
The target-preserving SMC path applies no top-$k$ or nucleus truncation, retaining
the support required by the importance correction.

Pass@1 averages four independently seeded complete executions with seeds
$\{0,1,2,3\}$. Pass@4 uses the same four executions. Coverage@32
uses the terminal particle set from one execution and is reported separately
because those candidates share ancestry. All timing measurements include image
processing, visual encoding, decoder prefill, token generation, resampling,
visual routing, and output selection.